\documentclass[11pt]{article}
\usepackage[preprint]{acl} 
\usepackage{times}
\usepackage{latexsym}
\usepackage[T1]{fontenc}
\usepackage[utf8]{inputenc}
\usepackage{microtype}
\usepackage{inconsolata}
\usepackage{graphicx}
\usepackage{booktabs}
\usepackage{adjustbox}
\usepackage{amsmath,amsfonts,amsthm}
\usepackage{nicefrac}
\usepackage{xcolor,colortbl}
\usepackage{multirow}
\usepackage{subcaption}
\usepackage{enumitem}
\usepackage{comment}
\usepackage{placeins}
\usepackage{pifont}
\newcommand{\papertablestyle}{%
  \footnotesize
  \setlength{\tabcolsep}{3.5pt}%
  \renewcommand{\arraystretch}{1.12}%
}

\newtheorem{proposition}{Proposition}
\newcommand{\cmark}{\checkmark}
\newcommand{\xmark}{\ding{55}}

\title{KiToke: Distribution-Aware Token Compression\\for Video Large Language Models}
\author{%
\makebox[\dimexpr\textwidth-2\tabcolsep\relax][c]{%
Haifeng Huang\textsuperscript{1}\qquad
Biyin Xu\textsuperscript{1}\qquad
Yang Li\textsuperscript{1}%
}%
}

\begin{document}
\maketitle
\begingroup
\renewcommand{\thefootnote}{\arabic{footnote}}
\footnotetext[1]{Iowa State University. Correspondence to: Yang Li (\href{mailto:jerryyangli@gmail.com}{jerryyangli@gmail.com}).}
\endgroup
\setcounter{footnote}{1}

\begin{abstract}
Video Large Language Models (Video LLMs) achieve strong performance on video understanding tasks but suffer from high inference costs due to the large number of visual tokens. We study training-free, query-agnostic video token compression, where heuristic importance rules can over-represent large redundant visual groups and miss less frequent visual patterns, particularly under aggressive token budgets. We propose KiToke, a distribution-aware compression framework that treats visual tokens as samples from the video's spatiotemporal token distribution. KiToke uses global embedding density as a proxy for representational redundancy and inverse-density sampling to balance token allocation across visual patterns of different frequencies. It further aggregates unselected tokens into retained anchors within detected temporal intervals to reduce cross-transition mixing. Across evaluated backbones, KiToke achieves the highest average accuracy among compared training-free methods and remains robust down to 1\% retention, where most evaluated baselines underperform uniform random token sampling. On LLaVA-OneVision, it preserves 98.5\% and 89.2\% of the uncompressed model's four-benchmark mean accuracy at 10\% and 1\% token retention, with 6.8$\times$ and 12.5$\times$ prefill speedups, respectively, including compression overhead.
\end{abstract}

\section{Introduction}

Video Large Language Models (Video LLMs)~\cite{qwen3-vl, internvl3.5} have achieved strong performance across a wide range of video understanding tasks, but their high inference cost remains a major barrier to practical deployment. A primary source of this cost is the large number of visual tokens extracted from videos. Videos contain substantial \emph{spatiotemporal redundancy}: repeated or highly similar content within frames and across time, such as static backgrounds, slow motion, and recurring scenes. This redundancy makes token compression a promising approach for reducing inference cost. In this work, we study \emph{training-free, query-agnostic} token compression, a plug-and-play setting in which visual tokens are compressed at inference time without retraining the model or relying on query-specific signals.

\begin{figure*}[t]
  \centering
    \includegraphics[width=0.99\linewidth]{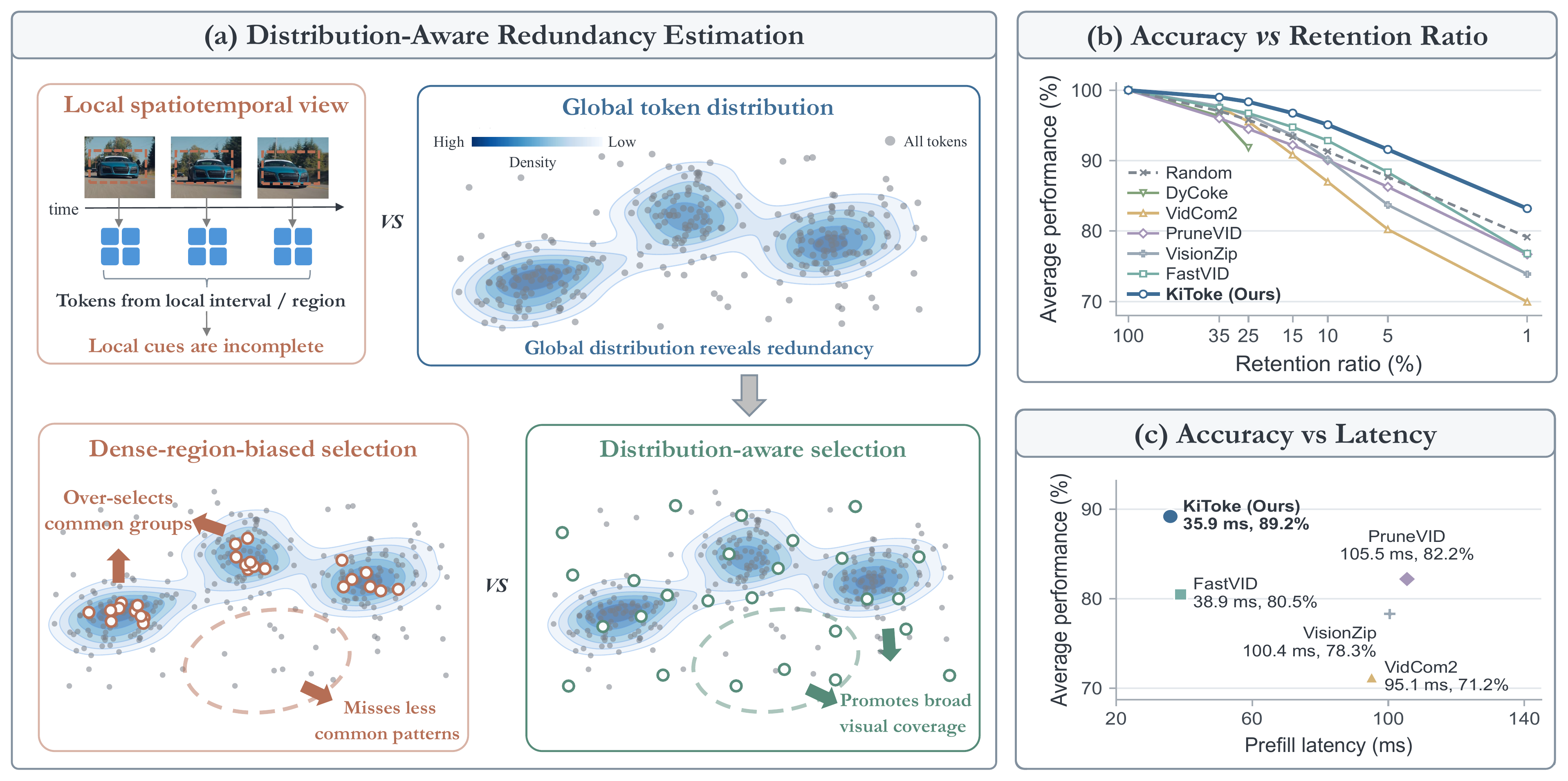}
    \caption{
        \textbf{Distribution-aware token compression improves robustness across retention ratios.}
        (a) KiToke uses global token density as a proxy for redundancy and inverse-density sampling to promote broad visual coverage.
        (b) Normalized performance vs.\ retention ratio for representative token compression methods, including Random. All prior methods shown at 1\% retention fall below Random. Performance is normalized by each uncompressed backbone before averaging over LLaVA-OneVision, LLaVA-Video, and Qwen3-VL, using four benchmarks.
        (c) Normalized average accuracy versus prefill latency on LLaVA-OneVision-7B at 1\% retention. Accuracy is the four-benchmark mean normalized to the uncompressed model. Latency includes compression overhead.
    }
    \label{fig:intro}
  
\end{figure*}

Recent training-free video token compression methods~\cite{prunevid, vidcom2, fastvid, aot, floc} exploit spatiotemporal redundancy through intuitive token selection criteria. VidCom2 treats tokens similar to an averaged visual token as redundant, while PruneVID assumes tokens that remain static across frames are more redundant than dynamic ones. FastVID selects representative density peaks within temporal intervals using local density and separation from higher-density tokens. While effective in some settings, these criteria capture only partial aspects of redundancy. A global-average criterion can be unreliable when tokens form multiple distant clusters, since a token near the average may still correspond to a distinct visual pattern. A static/dynamic distinction may miss redundant dynamic content that repeatedly appears across time. Similarly, selecting representatives within temporal intervals does not explicitly account for visual patterns recurring across distant intervals, potentially leaving less capacity for less frequent patterns. These limitations call for a more principled approach to quantifying token redundancy.

This need becomes especially pronounced under small token budgets, where repeated content can occupy a large fraction of the available tokens. As shown in Figure~\ref{fig:intro}(b), existing methods degrade sharply as the retention ratio decreases, with many underperforming a simple random token selection baseline at low retention ratios. These results motivate a selection rule that limits over-representation of repeated visual patterns under tight budgets.

Our objective is to preserve a broad range of visual content without predicting its relevance to a particular query. Conceptually, we view video token compression as a distribution-aware representation allocation problem, where the goal is to allocate a limited token budget across visual patterns with different frequencies. We assume that similar token embeddings tend to represent overlapping visual content. A token in a dense embedding region then has many potential substitutes, while one in a sparse region has fewer. We therefore measure redundancy through the prevalence of similar tokens across the full video.

We propose \textbf{KiToke}, a training-free, query-agnostic video token compression framework that realizes this distribution-level perspective. KiToke estimates redundancy from the full-video token distribution using kernel density estimation (KDE), then assigns sampling weights inversely proportional to this density. This full-video scope captures patterns that may be uncommon within individual frames or intervals yet recur throughout the video. In the idealized case of well-separated visual groups, our analysis in Section~\ref{sec:method_distribution} shows that inverse-density sampling assigns each group equal total sampling weight regardless of its size, thereby counterbalancing differences in pattern frequency.

To preserve additional visual information, KiToke merges unselected tokens into retained anchors using the same inverse-density scores, reducing the influence of repeated content on both anchor allocation and feature aggregation. While redundancy is estimated globally to account for repetition throughout the video, merging is restricted to content-adaptive temporal intervals to reduce cross-transition mixing. These intervals are constructed from token-level frame-to-frame changes, preserving local transition cues that averaged frame features may obscure.

We evaluate KiToke on multiple-choice, open-ended, and cross-domain video QA benchmarks using four Video LLM backbones at different scales. Across a wide range of retention ratios, KiToke achieves the strongest average performance among the evaluated training-free compression methods.

Our contributions are summarized as follows:
\begin{itemize}[leftmargin=*,nosep]
    \item To the best of our knowledge, this is the first work to formulate training-free, query-agnostic video token compression as distribution-aware token budget allocation that counterbalances visual-pattern frequency through full-video inverse-density weighting.
    \item We propose KiToke, a plug-and-play compressor that realizes this distribution-aware formulation by using global token redundancy to guide representation allocation, promoting coverage of less frequent visual patterns while preserving additional information through content-adaptive interval-aware merging.
    \item We demonstrate strong accuracy--efficiency trade-offs across benchmarks, backbones, and retention ratios. On LLaVA-OneVision, KiToke preserves 98.5\% and 89.2\% of the uncompressed model's average accuracy over four benchmarks at 10\% and 1\% token retention, with 6.8$\times$ and 12.5$\times$ prefill speedups, respectively, including compression overhead.
\end{itemize}

\section{Related Work}

\noindent\textbf{Video large language models.}
Building on multimodal LLMs~\cite{blip-2, llava-1.5, llava-next}, Video LLMs encode sampled frames as visual tokens for language-model reasoning. Representative designs include LLaVA-OneVision~\cite{llava-onevision}, LLaVA-Video~\cite{llava-video}, InternVL3.5~\cite{internvl3.5}, and the Qwen-VL series~\cite{qwen2.5-vl, qwen3-vl}, which strengthen spatiotemporal modeling through techniques like positional encoding and timestamp alignment. Understanding long videos still requires many frames and visual tokens, so the quadratic cost of attention~\cite{transformer} makes token compression an important step toward efficient inference.

\noindent\textbf{Token compression for video LLMs.}
Token compression improves efficiency by reducing the number of visual tokens processed by the LLM. While training-aware methods~\cite{pllava, llama-vid, apollo, longvu} incorporate compression into model training, we study training-free, query-agnostic approaches that operate at inference time and can be applied to different video LLMs, a setting widely used in prior work~\cite{aot, floc}.

Training-free methods initially focused on Image LLMs~\cite{pyramiddrop, vtw, divprune, llava-prumerge}, including query-aware pruning in FastV~\cite{fastv} and vision-encoder compression in VisionZip~\cite{visionzip}.

Recent work extends training-free compression to videos~\cite{dycoke, dytok, framefusion, mmg-vid, aot, floc} by accounting for spatiotemporal redundancy. VidCom2~\cite{vidcom2} measures proximity to the average visual token, PruneVID~\cite{prunevid} decouples temporal and spatial redundancy, and FastVID~\cite{fastvid} selects representative density peaks within temporal intervals using local density and separation from higher-density tokens.

Retaining many similar tokens can leave less room for other visual content, motivating selection methods that encourage diversity among the retained tokens. DivPrune favors pairwise diversity, while query-aware CDPruner~\cite{cdpruner} uses a determinantal point process for instruction-conditioned diversity. FLoC~\cite{floc} emphasizes coverage of the original sequence through a facility-location objective that sums each token's similarity to its closest representative. Because each occurrence contributes separately, repeated patterns can dominate this coverage objective even when they add little new visual information. KiToke instead uses full-video inverse-density weights to give less frequent visual patterns more opportunity to be selected, while limiting the influence of repeated content on merged features. The controlled comparisons in Section~\ref{sec:allocation_controls} show effectiveness of our approach when the merging is held fixed. Appendix~\ref{appendix:novelty} further discusses KiToke's conceptual and methodological novelty.

\section{Method}

\begin{figure*}[tp]
  \centering
  \includegraphics[width=\linewidth]{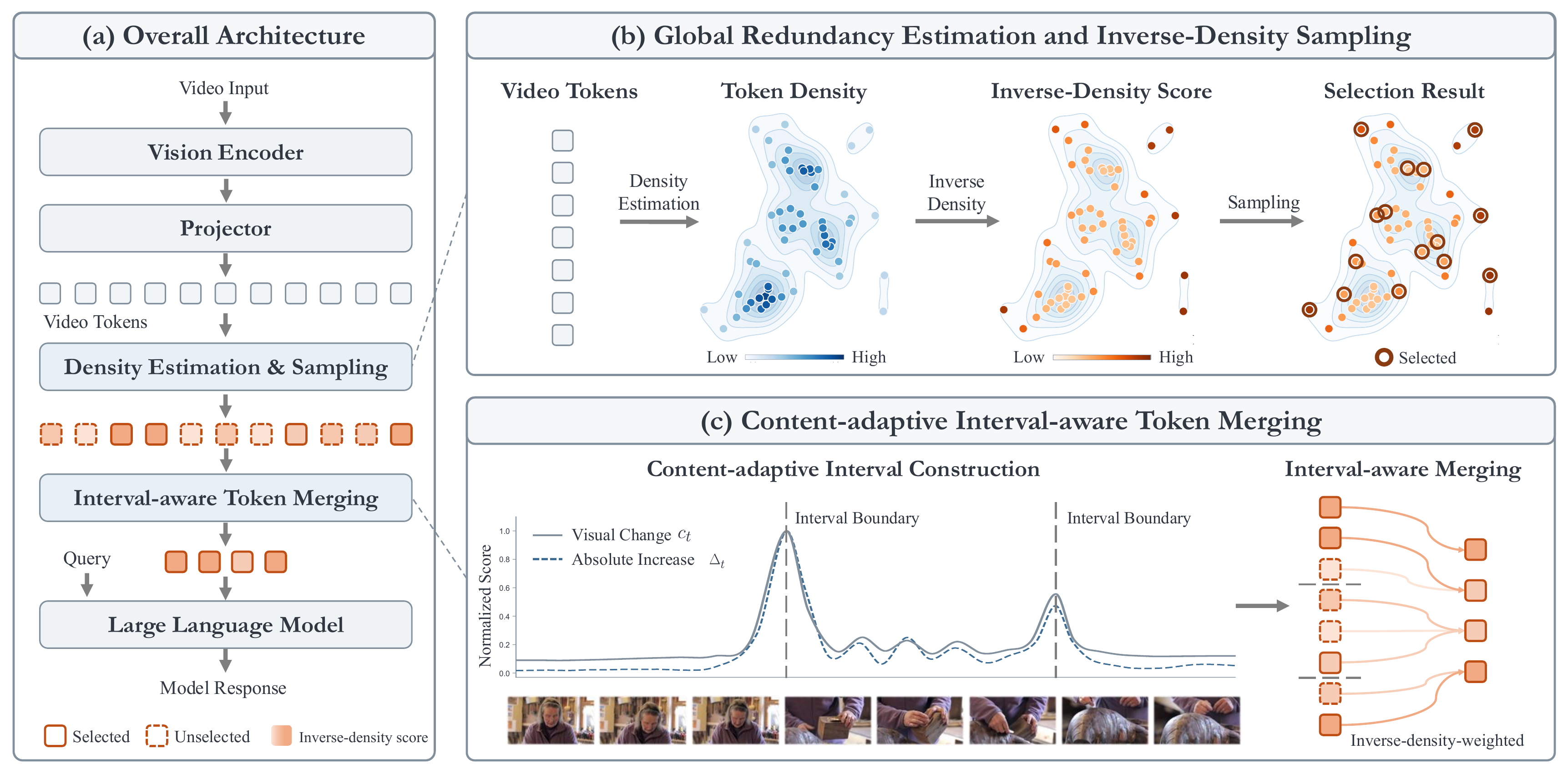}
  \caption{\textbf{Overview of KiToke.}
  (a) Visual tokens are compressed between the vision projector and the LLM.
  (b) Anchors are sampled using inverse-density scores computed from the full-video token distribution.
  (c) Unselected tokens are merged into anchors within content-adaptive temporal intervals using the same scores.}
  \label{fig:model}
\end{figure*}

\subsection{Overview}
Figure~\ref{fig:model} illustrates KiToke, a training-free, query-agnostic framework that compresses visual tokens before they enter the LLM. It first estimates redundancy from the full-video token distribution and samples anchors using inverse-density weights, counterbalancing differences in visual-pattern frequency. To preserve additional visual evidence, it then aggregates unselected tokens into these anchors within content-adaptive temporal intervals. Using the same inverse-density scores for selection and aggregation reduces the influence of repetition on both the anchor budget and the merged features.

\subsection{Distribution-aware Token Allocation\label{sec:method_distribution}}

In the query-agnostic setting, compression must preserve visual content without knowing which details a future query will require. We use embedding similarity as a proxy for overlapping visual content, so tokens with many similar neighbors have more potential substitutes. This motivates estimating redundancy across the full video and allocating less sampling weight to densely represented patterns.

\noindent\textbf{Estimating global pattern prevalence.}
Let $\{\mathbf{x}_i\}_{i=1}^N$ denote the visual token embeddings extracted from all frames of a video, where $N = T \times M$ for $T$ frames and $M$ tokens per frame. We estimate the density of this full-video token distribution using the Gaussian kernel:
\begin{equation}
K(\mathbf{x}_i, \mathbf{x}_j) = \exp\!\left(-\frac{\|\mathbf{x}_i - \mathbf{x}_j\|_2^2}{\alpha}\right),
\end{equation}
where $\alpha > 0$ controls how rapidly similarity decays with embedding distance. Summing these contributions over the full video gives the density at token $\mathbf{x}_i$:
\begin{equation}
D_i = \sum_{j=1}^N K(\mathbf{x}_i, \mathbf{x}_j).
\end{equation}
This sum is a Gaussian KDE up to a normalization constant. Higher density indicates more similar counterparts across the video, making it a proxy for representational redundancy. We therefore define the \emph{inverse-density score} $S_i=1/D_i$ to give greater weight to tokens with fewer potential substitutes.

We use full KDE by default in our experiments. Appendix~\ref{appendix:ann_kde} describes a variant based on approximate nearest-neighbor (ANN) search for more efficient compression of very long videos.

\noindent\textbf{Inverse-density sampling.}
Given a retention ratio $\gamma \in (0,1]$ and budget $B=\lfloor\gamma N\rfloor\geq1$, we assign each token the normalized sampling weight $q_i=S_i/\sum_{j=1}^N S_j$.

We first convert the weights into inclusion probabilities $\pi_i$, which specify each token's probability of being selected:
\begin{equation}\label{eq:inclusion_probabilities}
\pi_i=\min(1,\lambda q_i),\qquad \sum_{i=1}^N\pi_i=B.
\end{equation}
Here, $\lambda$ scales the weights so that the probabilities sum to $B$. Pivotal sampling~\cite{deville1998splitting} then selects exactly $B$ distinct anchors while preserving these individual selection probabilities. Appendix~\ref{appendix:fixed_size_sampling} describes the sampling procedure.

Under uniform sampling, patterns with more tokens receive more total sampling weight. Inverse-density weighting counterbalances this effect by reducing each occurrence's weight as similar tokens accumulate. The following proposition characterizes this behavior when density is approximately proportional to group size, allowing for imperfect within-group similarity and contributions from other groups.

\begin{samepage}
\begin{proposition}[Robust group allocation]\label{prop:group_allocation}
Partition the tokens into $G$ nonempty groups $\mathcal{G}_g$ of sizes $n_g$. Suppose $(1-\epsilon)n_g\leq D_i\leq(1+\eta)n_g$ for every $i\in\mathcal{G}_g$, where $0\leq\epsilon<1$ and $\eta\geq0$. With $Q_g=\sum_{i\in\mathcal{G}_g}q_i$ and $\rho=(1+\eta)/(1-\epsilon)$,
\begin{equation}\label{eq:group_allocation_bound}
\frac{1}{1+(G-1)\rho}\leq Q_g\leq\frac{\rho}{\rho+G-1}.
\end{equation}
\end{proposition}
\end{samepage}
When $\epsilon$ and $\eta$ are small, groups of different sizes receive nearly equal sampling mass, reaching exactly $1/G$ each at $\epsilon=\eta=0$. Under unsaturated inclusion probabilities $\pi_i=Bq_i$, the expected number of anchors sampled from group $g$ is $BQ_g$. Appendix~\ref{appendix:group_allocation} gives the proof and kernel conditions.

\subsection{Interval-aware Token Merging\label{sec:method_interval}}

Unselected tokens may still contain useful visual evidence, which we preserve by merging them into retained anchors. Although redundancy is estimated globally, we restrict aggregation to content-adaptive temporal intervals to reduce cross-transition mixing.

\noindent\textbf{Temporal intervals.}
Let $\{\mathcal{F}_t\}_{t=1}^T$ denote the video frames, where each frame $\mathcal{F}_t = \{\mathbf{v}_{t,i}\}_{i=1}^M$\footnote{Both $\mathbf{x}$ and $\mathbf{v}$ denote token embeddings with $\mathbf{x}_{M(t-1)+i} = \mathbf{v}_{t,i}$. We use $\mathbf{x}$ when tokens are indexed in a flattened sequence and $\mathbf{v}$ when frame and within-frame indices are explicit.} contains $M$ visual token embeddings. We measure change between consecutive frames using the token distance $d_t(i,j)=\|\mathbf{v}_{t-1,i}-\mathbf{v}_{t,j}\|_2$ and the frame-change score
\begin{equation}
c_t=\frac{1}{M}\sum_{i=1}^{M}\bigl[d_t(i,i)+\min_j d_t(i,j)\bigr].
\end{equation}
The first term compares tokens at corresponding spatial positions, while the second matches each token to its nearest counterpart in the next frame to account for displacement. Alongside the change magnitude $c_t$, we consider its largest increase over either neighboring transition, $\Delta_t=\max(c_t-c_{t-1},\,c_t-c_{t+1})$. We place a boundary before frame $t$ whenever $c_t>\tau_c$ or $\Delta_t>\tau_\Delta$.

\noindent\textbf{Merging strategy.}
With $L\leq B$ temporal intervals, we first repair any empty interval by adding its highest-score token and removing the lowest-score anchor from an interval with at least two anchors, as detailed in Appendix~\ref{appendix:empty_intervals}. Within interval $\mathcal{I}_k$, let $\mathcal{R}_k$ and $\mathcal{U}_k$ denote the anchors and unselected tokens after repair. Each $\mathbf{x}_u\in\mathcal{U}_k$ is assigned to its nearest anchor $m(\mathbf{x}_u)$ in $\mathcal{R}_k$ by L2 distance. For an anchor $\mathbf{x}_r$, let $\mathcal{A}_r$ contain its own index $r$ and the indices of all tokens assigned to it. The merged feature is their inverse-density-weighted average:
\begin{equation}
\tilde{\mathbf{x}}_r=\frac{\sum_{i\in\mathcal{A}_r} S_i\mathbf{x}_i}{\sum_{i\in\mathcal{A}_r} S_i}.
\end{equation}
Using the same scores as in sampling reduces the influence of repeated patterns on both anchor allocation and the merged features. We return the merged anchors $\{\tilde{\mathbf{x}}_r\}$ in their original sequence order, together with their anchor indices.

\section{Experiments}

\begin{table*}[!t]
\centering
\caption{\textbf{Performance comparison with state-of-the-art methods across video understanding benchmarks on \emph{LLaVA-OneVision}.} Best results under each retention ratio are highlighted in bold. Stochastic methods are averaged over 10 random seeds. Parentheses show differences from Random at the same retention ratio, in percentage points (pp) of relative accuracy. Benchmark accuracies are percentages, and relative accuracy is the four-benchmark mean divided by the uncompressed mean. Table~\ref{tab:comparison_llava_ov_full} reports all six retention ratios.}
\label{tab:comparison_llava_ov}
\papertablestyle
\begin{adjustbox}{max width=\linewidth}
\begin{tabular}{@{}l*{10}{c}@{}}
\toprule
\multirow{2}{*}{Method} & \multirow{2}{*}{\shortstack{Retention\\Ratio $\gamma$}} & \multirow{2}{*}{MVBench} & \multirow{2}{*}{LongVB} & \multirow{2}{*}{MLVU} & \multicolumn{4}{c}{VideoMME} & \multicolumn{2}{c@{}}{Mean accuracy} \\
\cmidrule(lr){6-9}\cmidrule(lr){10-11}
 &  &  &  &  & Overall & Short & Medium & Long & \shortstack{Absolute} & \multicolumn{1}{c@{}}{\shortstack{Relative}} \\
Duration
&  
& 16s
& 1$\sim$60min
& 3$\sim$120min
& 1$\sim$60min
& 1$\sim$3min
& 3$\sim$30min
& 30$\sim$60min
&  
&  
\\
\midrule
\textbf{LLaVA-OV-7B} & 100\% & 58.3 & 56.6 & 63.1 & 58.4 & 70.3 & 56.3 & 48.6 & 59.1 & 100.0\% \\
\midrule
\rowcolor{gray!6}
Random & 25\% & 56.7 & 57.1 & 61.9 & 56.7 & 68.6 & 54.4 & 47.2 & 58.1 & 98.3\% \\
FastV~\cite{fastv} & 100\%/25\% & 57.2 & 55.0 & 60.8 & 57.3 & 70.0 & 54.0 & 47.9 & 57.6 & 97.5\%~(\textcolor{red}{-0.8~pp}) \\
DyCoke~\cite{dycoke} & 25\% & 57.4 & 53.8 & 58.4 & 53.9 & 64.1 & 52.7 & 44.8 & 55.9 & 94.6\%~(\textcolor{red}{-3.7~pp}) \\
VidCom2~\cite{vidcom2} & 25\% & 57.0 & 56.3 & 62.4 & 57.4 & 68.1 & 56.0 & 48.2 & 58.3 & 98.6\%~(\textcolor{green!50!black}{+0.3~pp}) \\
PruneVID~\cite{prunevid} & 25\% & 56.8 & 55.6 & 62.8 & 56.9 & 68.7 & 54.1 & 48.0 & 58.0 & 98.1\%~(\textcolor{red}{-0.2~pp}) \\
VisionZip~\cite{visionzip} & 25\% & 57.7 & 56.0 & 62.1 & 57.7 & 69.6 & 56.0 & 47.5 & 58.4 & 98.8\%~(\textcolor{green!50!black}{+0.5~pp}) \\
AOT~\cite{aot} & 25\% & 58.2 & 55.8 & 62.8 & 57.6 & 69.0 & 55.2 & 48.7 & 58.6 & 99.2\%~(\textcolor{green!50!black}{+0.9~pp}) \\
FastVID~\cite{fastvid} & 25\% & 58.0 & 56.0 & 61.7 & 58.2 & \textbf{70.3} & \textbf{56.4} & 47.9 & 58.5 & 99.0\%~(\textcolor{green!50!black}{+0.7~pp}) \\
FLoC~\cite{floc} & 25\% & 56.7 & 56.8 & 62.1 & 57.0 & 68.8 & 54.9 & 47.3 & 58.2 & 98.5\%~(\textcolor{green!50!black}{+0.2~pp}) \\
\textbf{KiToke (Ours)} & 25\% & \textbf{58.4} & \textbf{57.4} & \textbf{63.4} & \textbf{58.3} & 70.0 & 56.1 & \textbf{48.8} & \textbf{59.4} & \textbf{100.5\%}~\textbf{(\textcolor{green!50!black}{+2.2~pp})} \\
\midrule
\rowcolor{gray!6}
Random & 10\% & 54.2 & 54.7 & 59.7 & 54.7 & 64.8 & 53.3 & 45.9 & 55.8 & 94.4\% \\
FastV~\cite{fastv} & 100\%/10\% & 53.0 & 50.7 & 57.1 & 53.6 & 63.0 & 51.9 & 46.0 & 53.6 & 90.7\%~(\textcolor{red}{-3.7~pp}) \\
VidCom2~\cite{vidcom2} & 10\% & 52.5 & 50.5 & 58.0 & 53.3 & 61.8 & 51.4 & 46.8 & 53.6 & 90.7\%~(\textcolor{red}{-3.7~pp}) \\
PruneVID~\cite{prunevid} & 10\% & 55.7 & 54.8 & 60.7 & 55.9 & \textbf{66.8} & 54.0 & 47.0 & 56.8 & 96.1\%~(\textcolor{green!50!black}{+1.7~pp}) \\
VisionZip~\cite{visionzip} & 10\% & 55.6 & 52.7 & 59.4 & 55.6 & 66.3 & 54.2 & 46.4 & 55.8 & 94.4\%~(0.0~pp) \\
AOT~\cite{aot} & 10\% & 56.9 & 53.0 & 59.3 & 54.1 & 64.6 & 51.3 & 46.4 & 55.8 & 94.4\%~(0.0~pp) \\
FastVID~\cite{fastvid} & 10\% & 56.6 & 55.6 & 60.4 & 55.6 & 66.2 & 54.2 & 46.4 & 57.0 & 96.4\%~(\textcolor{green!50!black}{+2.0~pp}) \\
FLoC~\cite{floc} & 10\% & 55.6 & 54.8 & 60.6 & 55.4 & 66.1 & 53.7 & 46.4 & 56.6 & 95.8\%~(\textcolor{green!50!black}{+1.4~pp}) \\
\textbf{KiToke (Ours)} & 10\% & \textbf{57.2} & \textbf{57.4} & \textbf{62.1} & \textbf{56.0} & 66.3 & \textbf{54.4} & \textbf{47.1} & \textbf{58.2} & \textbf{98.5\%}~\textbf{(\textcolor{green!50!black}{+4.1~pp})} \\
\midrule
\rowcolor{gray!6}
Random & 1\% & 48.0 & 46.9 & 54.6 & 49.0 & 54.4 & 49.1 & 43.3 & 49.6 & 83.9\% \\
FastV~\cite{fastv} & 100\%/1\% & 43.8 & 46.7 & 51.7 & 48.0 & 52.7 & 48.8 & 42.7 & 47.6 & 80.5\%~(\textcolor{red}{-3.4~pp}) \\
VidCom2~\cite{vidcom2} & 1\% & 39.2 & 41.7 & 45.7 & 41.7 & 42.0 & 42.1 & 40.9 & 42.1 & 71.2\%~(\textcolor{red}{-12.7~pp}) \\
PruneVID~\cite{prunevid} & 1\% & 46.0 & 46.7 & 53.7 & 48.0 & 53.0 & 47.7 & 43.3 & 48.6 & 82.2\%~(\textcolor{red}{-1.7~pp}) \\
VisionZip~\cite{visionzip} & 1\% & 42.6 & 43.9 & 52.1 & 46.5 & 49.6 & 47.3 & 42.7 & 46.3 & 78.3\%~(\textcolor{red}{-5.6~pp}) \\
AOT~\cite{aot} & 1\% & 45.5 & 44.4 & 51.1 & 46.1 & 49.2 & 46.1 & 43.1 & 46.8 & 79.2\%~(\textcolor{red}{-4.7~pp}) \\
FastVID~\cite{fastvid} & 1\% & 45.7 & 46.1 & 51.9 & 46.9 & 52.3 & 46.4 & 41.8 & 47.6 & 80.5\%~(\textcolor{red}{-3.4~pp}) \\
FLoC~\cite{floc} & 1\% & 47.3 & 47.0 & 51.4 & 47.7 & 54.8 & 47.0 & 41.3 & 48.4 & 81.9\%~(\textcolor{red}{-2.0~pp}) \\
\textbf{KiToke (Ours)} & 1\% & \textbf{52.5} & \textbf{51.1} & \textbf{56.9} & \textbf{50.2} & \textbf{57.4} & \textbf{49.3} & \textbf{43.8} & \textbf{52.7} & \textbf{89.2\%}~\textbf{(\textcolor{green!50!black}{+5.3~pp})} \\
\midrule
LLaVA-OV-7B & 0\% & 38.0 & 41.4 & 43.7 & 40.9 & 41.0 & 41.6 & 40.1 & 41.0 & 69.4\% \\
\bottomrule
\end{tabular}
\end{adjustbox}
\end{table*}

\subsection{Experimental Settings}

\noindent\textbf{Benchmarks.}
We evaluate on four widely-adopted VQA benchmarks: MVBench~\cite{mvbench}, LongVideoBench~\cite{longvideobench}, abbreviated as LongVB in tables, MLVU~\cite{mlvu}, and VideoMME without subtitles~\cite{video-mme}. We also assess transfer to open-ended QA with CVRR-ES~\cite{cvrr-es} and VideoEval-Pro~\cite{videoeval-pro}, and to cross-domain tasks with EgoSchema~\cite{egoschema} and Video-MMMU~\cite{video-mmmu}.

\noindent\textbf{Implementation details.}
We evaluate four representative video LLMs: LLaVA-OneVision~\cite{llava-onevision}, LLaVA-Video~\cite{llava-video}, Qwen3-VL~\cite{qwen3-vl}, and InternVL3.5~\cite{internvl3.5} using LMMs-Eval~\cite{lmms-eval} on NVIDIA A100 GPUs. We calibrate the Gaussian bandwidth and interval thresholds once from unlabeled LLaVA-OneVision features, obtaining $\alpha{=}800$ and thresholds $(110,70)$ on $(c_t,\Delta_t)$. Unless otherwise stated, these settings are fixed across backbones and benchmarks. Appendices~\ref{appendix:backbone_settings} and~\ref{appendix:hyperparameter_calibration} describe the backbone settings and calibration procedure.

\begin{figure*}[tp]
  \centering
  \begin{subfigure}{0.48\textwidth}
    \centering
    \includegraphics[width=\linewidth]{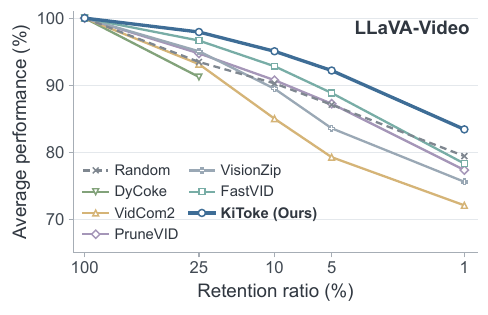}
  \end{subfigure}
  \hfill
  \begin{subfigure}{0.48\textwidth}
    \centering
    \includegraphics[width=\linewidth]{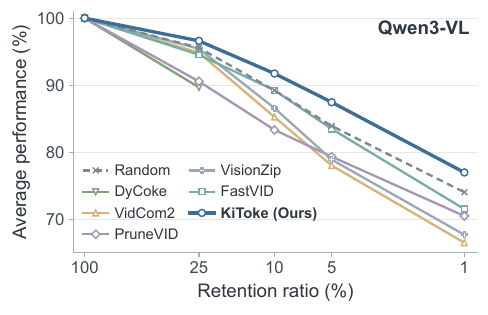}
  \end{subfigure}
  \caption{
    \textbf{Performance comparison with state-of-the-art methods across video understanding benchmarks on \emph{LLaVA-Video} and \emph{Qwen3-VL}.}
    Mean accuracy over four video understanding benchmarks is normalized to the uncompressed backbone (100\%). Detailed benchmark-level results are reported in Appendix~\ref{appendix:detailed_results}.
  }
  \label{fig:additional_backbone_retention}
\end{figure*}

\begin{table*}[tp]
  \centering
  \caption{\textbf{Ablation study of core components.} ``Density Est.'' denotes global redundancy estimation, while ``Interval'' and ``Global'' denote interval-based and full-sequence token merging. Row 3 uses uniform merge weights. Parentheses report relative accuracy (\%), normalized by the uncompressed mean.}
  \label{tab:token_merge_ablation}
  \papertablestyle
\begin{adjustbox}{max width=\linewidth}
  \begin{tabular}{@{}*{9}{c}@{}}
  \toprule
  Row & Density Est. & Token Selection & Token Merge & MVBench & LongVB & MLVU & VideoMME & Mean accuracy (\%) \\
  \midrule
  1 & \xmark & Sampling & \xmark & 48.0 & 46.9 & 54.6 & 49.0 & 49.6 (83.9\%) \\
  2 & \cmark & Sampling & \xmark & 51.5 & 49.1 & 54.9 & 48.9 & 51.1 (86.5\%) \\
  3 & \xmark & Sampling & Interval & 50.1 & 47.8 & 54.0 & 49.3 & 50.3 (85.1\%) \\
  4 & \cmark & Top-K & Interval & 51.1 & 48.9 & 55.5 & 49.1 & 51.2 (86.6\%) \\
  5 & \cmark & Sampling & Global & 51.5 & 49.7 & 56.6 & 49.4 & 51.8 (87.6\%)\\
  6 & \cmark & Sampling & Interval & \textbf{52.5} & \textbf{51.1} & \textbf{56.9} & \textbf{50.2} & \textbf{52.7 (89.2\%)} \\
  \bottomrule
  \end{tabular}
\end{adjustbox}
  \end{table*}

\subsection{Main Comparisons}
\label{sec:low_retention_failure}

We compare KiToke with prior methods and uniform random token selection at matched retention ratios, under the unified evaluation protocol described in Appendix~\ref{appendix:exp_detail}. Stochastic results are averaged over 10 seeds, and differences below are reported in percentage points of four-benchmark mean accuracy.

\noindent\textbf{Results on LLaVA-OneVision.}
KiToke achieves the highest mean accuracy at every retention ratio in Table~\ref{tab:comparison_llava_ov}. At 25\% retention, Random and the strongest prior method already preserve 98.3\% and 99.2\% of uncompressed accuracy, respectively, yet KiToke reaches 100.5\%. Thus, its allocation strategy remains effective even when the budget can accommodate considerable redundancy. At 1\% retention, KiToke reaches 52.7\% accuracy, compared with 49.6\% for Random and 41.0\% for the language-only baseline.

\noindent\textbf{Results on LLaVA-Video and Qwen3-VL.}
Figure~\ref{fig:additional_backbone_retention} shows that KiToke also achieves the highest average accuracy on LLaVA-Video and Qwen3-VL. These gains extend to models with different spatiotemporal representations: LLaVA-Video uses newline tokens, while Qwen3-VL uses timestamp encoding with Interleaved-MRoPE.

\noindent\textbf{Preserving information at low retention.}
Low retention provides a stringent test of compression quality: large token budgets can tolerate redundant selections, whereas small budgets make each retained token more consequential. At 1\% retention, all evaluated prior methods fall below Random in mean accuracy on the three backbones in Table~\ref{tab:comparison_llava_ov} and Figure~\ref{fig:additional_backbone_retention}, showing that their compression strategies preserve useful information less effectively than random sampling at this budget. Even the strongest prior results fall below the lower bounds of Random's 95\% confidence intervals across 10 seeds in Table~\ref{tab:stochastic_ci_template}. KiToke instead outperforms Random on all three backbones, demonstrating the value of distribution-aware compression when the budget leaves little room for redundancy.

\noindent\textbf{Visual-encoder coverage.}
Our four backbones span three visual-encoder types: SigLIP~\cite{siglip} in both LLaVA models, a SigLIP-2-based encoder~\cite{siglip2, qwen3-vl} in Qwen3-VL, and InternViT~\cite{internvit} in InternVL3.5. Table~\ref{tab:comparison_internvl} shows that KiToke outperforms VidCom2 and PruneVID on InternVL3.5 at both 10\% and 1\% retention without retuning.

\noindent\textbf{Open-ended and cross-domain evaluation.}
In Table~\ref{tab:open_ended_cross_domain}, KiToke outperforms FastVID and PruneVID on all four transfer benchmarks without retuning. At 1\% retention, it achieves 58.2 on EgoSchema and 47.7 on CVRR-ES, compared with 48.8 and 41.6 for PruneVID.

\subsection{Ablation Study}
\label{sec:allocation_controls}
Unless otherwise stated, ablations use LLaVA-OneVision at 1\% retention and report the four-benchmark average.

\noindent\textbf{Core components.}
Table~\ref{tab:token_merge_ablation} examines how aggregation complements selection. With inverse-density sampling fixed, mean accuracy rises from 51.1\% without merging to 51.8\% with global merging and 52.7\% with interval-aware merging. The controls in Table~\ref{tab:temporal_merge_ablation} further separate the effects of boundary placement and merge weighting: KiToke gains 1.3 points over uniform intervals with matched counts, while inverse-density weights add 0.5 points with identical retained anchors and assignments.

\begin{table}[tp]
\centering
\caption{\textbf{Controlled query-agnostic token allocation.} We vary density scope only for selection, keeping intervals and inverse-density merge weights fixed for all selectors. Four-benchmark mean accuracy (\%) on LLaVA-OneVision-7B at 1\% retention is reported.}
\label{tab:allocation_principles}
\papertablestyle
\begin{tabular}{@{}lc@{}}
\toprule
Anchor selector & Mean accuracy (\%) \\
\midrule
Uniform sampling & 50.4 \\
Density-proportional, full video & 48.8 \\
\midrule
$k$-center~\cite{k-center} & 50.1 \\
DivPrune~\cite{divprune} & 50.8 \\
FastVID~\cite{fastvid} & 50.4 \\
FLoC~\cite{floc} & 50.3 \\
\midrule
Inverse density, per frame & 50.8 \\
Inverse density, per interval & 51.5 \\
\textbf{Inverse density, full video (ours)} & \textbf{52.7} \\
\bottomrule
\end{tabular}
\end{table}

\noindent\textbf{Controlled token allocation.}
To isolate allocation, Table~\ref{tab:allocation_principles} compares selectors with identical inputs, budgets, intervals and merging rules. Sampling variants share the pivotal sampler, with benchmark-level results and details provided in Appendix~\ref{appendix:selection_fixed_merging}.

With aggregation held fixed, full-video inverse-density sampling outperforms uniform and density-proportional sampling, as well as pairwise-diversity and facility-location selection.

\begin{table*}[tp]
\centering
\caption{\textbf{Efficiency comparison on LLaVA-OneVision.} LLM compute is reported in TFLOPs. Token counts are averaged over evaluation samples. Relative accuracy is normalized by the uncompressed mean. Total prefill latency includes compression overhead. Compression and LLM prefill times are averaged over 100 VideoMME samples at batch size 1 on a single NVIDIA A100-80GB GPU.}
\label{tab:efficiency}
\papertablestyle
\setlength{\tabcolsep}{4pt}
\begin{adjustbox}{max width=\linewidth}
\begin{tabular}{@{}lccccccccc@{}}
\toprule
\multirow{2}{*}{Method} & \multirow{2}{*}{\shortstack{Retention\\Ratio}} & \multirow{2}{*}{\shortstack{Mean \#\\visual tokens}} & \multirow{2}{*}{TFLOPs} & \multicolumn{3}{c}{Latency (ms)} & \multirow{2}{*}{Speedup} & \multicolumn{2}{c}{Mean accuracy} \\
\cmidrule(lr){5-7}\cmidrule(lr){9-10}
 & & & & Comp. & LLM & Total & & \shortstack{Absolute} & \shortstack{Relative} \\
\midrule
Vanilla & 100\% & 6272 & 97.64 & - & 448.0 & 448.0 & (1.0x) & 59.1 & 100\% \\
\midrule
VidCom2~\cite{vidcom2} & 10\% & 627.0 & 8.34 & 70.7 & 53.1 & 123.8 & (3.6x) & 53.6 & 90.7\% \\
PruneVID~\cite{prunevid} & 10\% & 628.8 & 8.36 & 81.2 & 53.9 & 135.1 & (3.3x) & 56.8 & 96.1\% \\
VisionZip~\cite{visionzip} & 10\% & 640.0 & 8.52 & 76.4 & 56.5 & 132.9 & (3.4x) & 55.8 & 94.4\% \\
FastVID~\cite{fastvid} & 10\% & 639.8 & 8.52 & 15.3 & 56.1 & 71.4 & (6.3x) & 57.0 & 96.4\% \\
\textbf{KiToke (Ours)} & 10\% & \textbf{627.0} & \textbf{8.34} & \textbf{13.1} & \textbf{53.1} & \textbf{66.2} & \textbf{(6.8x)} & \textbf{58.2} & \textbf{98.5\%} \\
\midrule
VidCom2~\cite{vidcom2} & 1\% & 65.5 & 0.86 & 70.8 & 24.3 & 95.1 & (4.7x) & 42.1 & 71.2\% \\
PruneVID~\cite{prunevid} & 1\% & 67.2 & 0.88 & 81.0 & 24.5 & 105.5 & (4.2x) & 48.6 & 82.2\% \\
VisionZip~\cite{visionzip} & 1\% & 64.0 & 0.84 & 76.5 & 23.9 & 100.4 & (4.5x) & 46.3 & 78.3\% \\
FastVID~\cite{fastvid} & 1\% & 62.9 & 0.82 & 15.2 & 23.7 & 38.9 & (11.5x) & 47.6 & 80.5\% \\
\textbf{KiToke (Ours)} & 1\% & \textbf{62.0} & \textbf{0.81} & \textbf{12.3} & \textbf{23.6} & \textbf{35.9} & \textbf{(12.5x)} & \textbf{52.7} & \textbf{89.2\%} \\
\bottomrule
\end{tabular}
\end{adjustbox}
\end{table*}

\noindent\textbf{Density-estimation scope.}
Restricting density estimation to individual frames or intervals reduces mean accuracy from 52.7\% to 50.8\% or 51.5\%, respectively. Full-video estimation improves all four benchmarks, highlighting the value of accounting for visual patterns that recur in different parts of the video.

\noindent\textbf{Sampling and diversity-based alternatives.}
With inverse-density scores fixed, Table~\ref{tab:token_selection_ablation} shows that pivotal sampling outperforms multinomial sampling and top-$K$. On LLaVA-Video, KiToke also outperforms $k$-center, DivPrune, and CDPruner at all three token budgets in Appendix~\ref{appendix:comparison_diversity}. Its lead over the strongest baseline grows from 0.9 to 3.7 accuracy percentage points as the budget decreases from 4096 to 1024 tokens.

\noindent\textbf{Robustness and transfer.}
Appendices~\ref{appendix:hyperparameter_sensitivity} and~\ref{appendix:model_size} show that KiToke's performance varies smoothly with moderate changes to the bandwidth and interval thresholds, and its gains extend to 0.5B and 72B models. Appendix~\ref{appendix:kernel_sensitivity} shows that Gaussian, Laplacian, and angular Gaussian variants differ by at most 0.4 percentage points in mean accuracy, suggesting robustness across these kernels.

\noindent\textbf{Query-aware pruning and saliency.}
At 5\% final retention, Table~\ref{tab:query_composition_full} shows that KiToke reaches 56.2\% mean accuracy, outperforming query-aware FastV, full PruneVID, and FastVID + FastV. Under the same 10\%-to-5\% schedule, KiToke + FastV reaches 56.8\%, versus 54.2\% for FastVID + FastV. Separately, Appendix~\ref{appendix:global_attention_prior} shows that a frame-global attention prior improves VideoMME accuracy by 1.0 point at 1\% retention.

\subsection{Efficiency and Scalability}

\noindent\textbf{Efficiency comparison.}
We report prefill latency including compression overhead in Table~\ref{tab:efficiency}, covering token compression and LLM prefill. KiToke has the lowest compression overhead and highest mean accuracy among the compared methods, with speedups over the uncompressed model of 6.8$\times$ at 10\% retention and 12.5$\times$ at 1\%. At 1\%, it improves mean accuracy over FastVID from 47.6\% to 52.7\% while reducing prefill latency from 38.9 to 35.9\,ms. Timing includes density estimation but excludes video decoding, vision encoding, and autoregressive decoding, as specified in Appendix~\ref{appendix:latency_protocol}. Appendix~\ref{appendix:efficiency_qwen3_vl} extends the comparison to 16K/32K-token Qwen3-VL inputs.

\noindent\textbf{Fixed-budget long-video inference.}
Processing more frames under a fixed output budget lowers the retention ratio while expanding temporal coverage. In Appendix~\ref{sec:fixed_budget}, increasing LLaVA-OneVision's input from 32 to 320 frames improves VideoMME accuracy from 58.4 to 61.6 with the LLM input fixed at 6272 tokens.

\noindent\textbf{Approximate density estimation.}
As an optional alternative to full KDE, KiToke's ANN variant avoids constructing the full pairwise kernel matrix. On Qwen3-VL-8B with 32K candidate tokens, the results in Appendix~\ref{appendix:ann_kde} show 8.5$\times$ lower density-estimation time and 24.8$\times$ lower incremental peak GPU allocation. At 1\% retention, prefill latency falls from 0.87 to 0.42\,s, while mean accuracy changes from 54.1\% to 53.9\%, offering a practical accuracy--cost trade-off.

\section{Conclusion}

KiToke formulates training-free, query-agnostic video token compression as distribution-aware budget allocation. It samples tokens using full-video inverse-density scores to balance allocation across visual patterns of different frequencies. To preserve information from unselected tokens, it then merges them into retained anchors within content-adaptive temporal intervals, using the same scores to weight their contributions. Across four Video LLM backbones and a wide range of retention ratios, KiToke achieves the highest average accuracy among the evaluated training-free methods. On LLaVA-OneVision, it preserves 98.5\% and 89.2\% of the uncompressed model's four-benchmark mean accuracy at 10\% and 1\% token retention, with 6.8$\times$ and 12.5$\times$ prefill speedups, respectively, including compression overhead.

\section*{Limitations}\label{sec:limitations}
\noindent\textbf{Compatibility with specialized spatiotemporal model designs.}
At low retention ratios, KiToke loses more accuracy relative to the uncompressed model on LLaVA-Video~\cite{llava-video} and Qwen3-VL~\cite{qwen3-vl} than on LLaVA-OneVision~\cite{llava-onevision}. Differences in spatiotemporal representations may contribute to this variation: LLaVA-Video uses newline tokens, while Qwen3-VL uses timestamp encoding with Interleaved-MRoPE. Understanding how compression interacts with these structures could guide adaptations to different model designs.

\noindent\textbf{Visual coverage and question-specific relevance.}
In the query-agnostic setting, compression is performed without access to the question, so it cannot directly identify which content will be useful for answering it. KiToke therefore allocates the budget to preserve broad visual coverage by counterbalancing pattern frequency. Our group-allocation analysis provides a rationale for this principle, while the benchmark results demonstrate its empirical effectiveness. The compressed representations can be computed offline and reused across questions or conversation turns. Adapting allocation to a particular question, such as favoring repeated occurrences for counting or distinct events for summarization, requires query-aware selection. This can be applied directly or combined with query-agnostic compression. Appendix~\ref{appendix:query_composition} evaluates KiToke followed by FastV, showing how broad visual coverage can complement question-specific selection.

\noindent\textbf{Evaluation of retained visual evidence.}
We evaluate compression quality through downstream video question answering, covering both multiple-choice and open-ended tasks. These answer-level metrics assess performance on the evaluated questions but do not directly measure which visual evidence remains recoverable after selection and merging. Developing a dedicated benchmark with evidence-level annotations and a protocol for assessing information preservation in compressed representations is an important direction for future research.

\begingroup
\renewcommand{\href}[2]{#2}
\bibliography{main}
\endgroup

\clearpage
\appendix
The appendix provides the experimental details, supporting results, and further analysis of KiToke:
\begin{enumerate}
    \item \textbf{Appendix~\ref{appendix:exp_detail}: Experimental details.} We describe the evaluation protocols, hyperparameter calibration, sampling and merging procedures, baseline implementations, and computational-cost measurements.
    \item \textbf{Appendix~\ref{appendix:detailed_results}: Detailed benchmark results.} We report performance across four Video LLM backbones and provide the evaluation protocols and results for open-ended and cross-domain tasks.
    \item \textbf{Appendix~\ref{appendix:exp_results}: Additional experiments and ablations.} Controlled comparisons examine token selection, full-video density estimation, temporal boundaries, and merge weighting. Further experiments assess scalability, sampling stability, sensitivity to design choices, and composition with query-aware pruning. The selector and temporal controls appear in Appendices~\ref{appendix:selection_fixed_merging}--\ref{appendix:temporal_merge_ablation}, and approximate density estimation is evaluated in Appendix~\ref{appendix:ann_kde}.
    \item \textbf{Appendix~\ref{appendix:limitations}: Additional discussion and theoretical analysis.} We explain how KiToke's allocation and aggregation rules differ from prior approaches and discuss its relationship to query-aware compression. Appendix~\ref{appendix:group_allocation} provides the proof and interpretation of Proposition~\ref{prop:group_allocation}.
    \item \textbf{Appendix~\ref{appendix:qualitative}: Qualitative analysis.} We visualize temporal boundaries and selected tokens, relating retained visual content to model predictions through successful and unsuccessful examples.
\end{enumerate}

\newcommand{\appendixanalysis}[1]{%
  \par\medskip
  \begin{minipage}{\textwidth}
    \normalfont\normalsize\noindent #1\par
  \end{minipage}%
}


\textbf{\textit{The code will be released when the paper is published.}}

\section{Experimental Details\label{appendix:exp_detail}}

\subsection{Evaluation and Backbone Settings}\label{appendix:backbone_settings}

Our evaluation covers four multiple-choice benchmarks: MVBench, LongVideoBench, MLVU, and VideoMME without subtitles. We also evaluate open-ended QA on CVRR-ES and VideoEval-Pro, and cross-domain transfer on EgoSchema and Video-MMMU.

The three main backbones use different frame-sampling and tokenization settings. LLaVA-OneVision samples 32 frames ($32{\times}196$ visual tokens), LLaVA-Video samples 64 frames ($64{\times}169$ tokens), and Qwen3-VL uses $\texttt{MAX\_PIXELS}=262{,}144$ with up to 128 frames and up to $16{,}384$ tokens. Unless otherwise stated, all main comparisons use Gaussian bandwidth $\alpha{=}800$ and interval-construction thresholds $\tau_c{=}110$ and $\tau_\Delta{=}70$ on $c_t$ and $\Delta_t$. These defaults are obtained through the calibration procedure in Appendix~\ref{appendix:hyperparameter_calibration}.

All experiments use LMMs-Eval\footnote{\url{https://github.com/EvolvingLMMs-Lab/lmms-eval}; MIT.}~\cite{lmms-eval} with the same software, hardware, and backbone checkpoints for all methods. We use the official implementations of LLaVA-OneVision\footnote{\label{fn:llava-next}\url{https://github.com/LLaVA-VL/LLaVA-NeXT}; Apache-2.0.}~\cite{llava-onevision}, LLaVA-Video\footref{fn:llava-next}~\cite{llava-video}, and Qwen3-VL\footnote{\url{https://github.com/QwenLM/Qwen3-VL}; Apache-2.0.}~\cite{qwen3-vl}.

\subsection{Hyperparameter Calibration}\label{appendix:hyperparameter_calibration}

We calibrate the Gaussian bandwidth and temporal thresholds on LLaVA-OneVision using 100 randomly sampled videos from the MLVU test split. Calibration uses only unlabeled visual-feature statistics, without questions, answers, or benchmark accuracy. All reported MLVU results are evaluated on the disjoint MLVU dev split, following common evaluation practice in prior work.

\noindent\textbf{Gaussian bandwidth.}
We extract the token embeddings used for redundancy estimation with the usual preprocessing, then uniformly sample 10,000 off-diagonal token pairs from each video's full sequence. Pooling their squared Euclidean distances into $\mathcal{D}$, we use the first percentile as a fixed heuristic for the bandwidth:
\[
\alpha_{\mathrm{cal}}=Q_{0.01}(\mathcal{D}),
\]
where $Q_{0.01}$ denotes the first percentile, setting the similarity scale from close token pairs. This gives $\alpha_{\mathrm{cal}}=827$, which we round to the nearest hundred to obtain the default $\alpha=800$.

\noindent\textbf{Temporal thresholds.}
For each calibration video, thresholds are the 80th percentiles of the empirical $c_t$ and $\Delta_t$ distributions, yielding \mbox{$(\tau_c,\tau_\Delta)=(110,70)$}.

Unless otherwise stated, we keep the calibrated defaults $\alpha=800$, $\tau_c=110$, and $\tau_\Delta=70$ fixed across all evaluated backbones, benchmarks, and retention ratios. Tables~\ref{tab:comparison_llava_ov_full}--\ref{tab:comparison_internvl} show that the gains transfer across LLaVA-OneVision, LLaVA-Video, Qwen3-VL, and InternVL3.5 despite their different visual encoders and spatiotemporal tokenization. Appendix~\ref{appendix:hyperparameter_sensitivity} examines how performance varies around these defaults.

\subsection{KiToke Implementation}

\subsubsection{Temporal Interval Construction}
For each candidate boundary before frame $t$, where $2\le t\le T$, we compute $c_t$ from frames $t-1$ and $t$ as in Section~\ref{sec:method_interval}. Inserting this boundary ends the current interval at frame $t-1$ and starts the next at frame $t$. For interior candidates $2<t<T$, $\Delta_t$ measures the largest increase relative to either neighboring transition score, $c_{t-1}$ or $c_{t+1}$. At the first and last candidate boundaries, we replace the missing side with the available side. For videos containing only two frames, we set $\Delta_2=0$. We insert a boundary whenever $c_t>\tau_c$ or $\Delta_t>\tau_\Delta$, so the absolute threshold on $c_t$ alone determines the boundary in the two-frame case.

\subsubsection{Fixed-size Sampling}\label{appendix:fixed_size_sampling}
Starting from $Bq_i$, the sampler fixes probabilities that reach one and repeatedly rescales the remaining probabilities to match the remaining budget. It then applies pivotal sampling to the fractional probabilities: each round uniformly shuffles the candidate tokens and pairs consecutive entries, carrying an unmatched token to the next round. The standard pivotal updates~\cite{deville1998splitting} preserve the specified inclusion probabilities and yield exactly $B$ anchors in exact arithmetic.

\subsubsection{Budget Feasibility and Empty Intervals}\label{appendix:empty_intervals}
Representing every interval requires at least as many anchors as intervals. For a budget $1\leq B\leq N$, if the initial interval count $L$ exceeds $B$, we therefore remove the $L-B$ boundaries with the smallest transition scores $c_t$, breaking ties chronologically. This merges adjacent intervals before sampling. After sampling, we add the highest-score token from each empty interval, then remove the lowest-score retained tokens from intervals containing at least two anchors until the count returns to $B$. The repair preserves the total budget and ensures interval coverage, while changing individual inclusion probabilities. All evaluated inputs satisfy $L\leq B$, so the coarsening fallback is inactive. Appendix~\ref{appendix:repair_ablation} examines the effect of the repair step.

\subsubsection{Token Merging}
For merging, each unselected token $\mathbf{x}_u$ is assigned to the selected anchor $\mathbf{x}_r$ with minimum L2 distance within the same interval. With the default weighted-average merge used in the experiments, the resulting anchor is
\[
\tilde{\mathbf{x}}_r
=
\frac{
S_r\mathbf{x}_r + \sum_{\mathbf{x}_u:m(\mathbf{x}_u)=\mathbf{x}_r} S_u\mathbf{x}_u
}{
S_r + \sum_{\mathbf{x}_u:m(\mathbf{x}_u)=\mathbf{x}_r} S_u
}.
\]
The anchor itself contributes with weight $S_r$, so it remains unchanged when no unselected tokens are assigned to it.

\subsubsection{Output Order and Positional Handling}\label{appendix:newline_tokens}
Merged anchors retain their original token indices and are returned in frame and within-frame order. The backbones then handle LLM positions according to their input conventions: LLaVA assigns consecutive position IDs after compression, whereas Qwen3-VL retains the positional IDs at the selected positions in the multimodal input.

To better preserve spatiotemporal structure, LLaVA-Video inserts a newline token after each row of 13 spatial tokens, adding 832 newline tokens on top of the original $64\times13\times13=10816$ visual tokens. Following FastVID, we retain a newline token if and only if at least one associated visual token is retained.

\subsection{Baseline Reproduction}

We adapt the baseline interfaces where needed to accommodate the backbone input/output formats and token-accounting rules. The following settings specify the compression stages and configurations used in our comparisons.

\noindent\textbf{FastV\footnote{\url{https://github.com/pkunlp-icler/FastV}.}~\cite{fastv} (ECCV 2024).} FastV performs token pruning at the $K$-th transformer layer based on attention scores, controlled by a filtering ratio $R$. In our experiments, we set $K=10$ and $R=1-\gamma$, where $\gamma$ is the retention ratio.

\noindent\textbf{DyCoke\footnote{\url{https://github.com/KD-TAO/DyCoke}; Apache-2.0.}~\cite{dycoke} (CVPR 2025).} DyCoke performs token pruning during both the prefill and decode stages. We evaluate its prefill-stage TTM module, which groups frames in fours and retains all visual tokens from the first frame. For each of the other three frames, we set the retention fraction to $r=\frac{4}{3}(\gamma-0.25)$, corresponding to \texttt{merging\_ratio} in our implementation. Before integer rounding, the overall visual-token retention is $\gamma=\frac14+\frac34r$. Thus, the minimum is $\gamma=25\%$: $r=0$ and only the first frame's tokens remain.

\noindent\textbf{VidCom2\footnote{\url{https://github.com/xuyang-liu16/VidCom2}; Apache-2.0.}~\cite{vidcom2} (EMNLP 2025).} VidCom2 prunes tokens by comparing each token to frame-level and video-level mean features. We follow their original implementation, setting the token scoring mode to $\texttt{negative\_video\_mean\_and\_global\_mean}$ and the frame scoring mode to $\texttt{negative\_video\_mean}$.

\noindent\textbf{PruneVID\footnote{\url{https://github.com/Visual-AI/PruneVid}; CC BY-NC-SA 4.0.}~\cite{prunevid} (ACL 2025).} PruneVID prunes tokens both before and within the LLM, while our main comparison evaluates its pre-LLM module. We use the official threshold of $0.8$, temporal segment ratio of $0.25$, and token selection ratio of $0.4$, while setting the cluster ratio $\beta$ to match each target retention ratio $\gamma$ under the token-budget accounting rule shared by all methods.

\noindent\textbf{VisionZip\footnote{\url{https://github.com/JIA-Lab-research/VisionZip}; Apache-2.0.}~\cite{visionzip} (CVPR 2025).} VisionZip was designed for image-level token pruning. Following prior Video LLM compression comparisons~\cite{fastvid}, we apply it after pooling, frame by frame, with the official dominant/context split ($85{:}15$), while preserving original token indices and temporal order.

\noindent\textbf{FastVID\footnote{\url{https://github.com/LunarShen/FastVID}; MIT.}~\cite{fastvid} (NeurIPS 2025).} We use FastVID's official pre-LLM implementation with its default hyperparameters: $c=8$, $\tau=0.9$, $d=0.4$, $p=4$, and $\beta=0.6$.

\noindent\textbf{AOT\footnote{\url{https://github.com/TyroneLi/AOT}; Apache-2.0.}~\cite{aot} (CVPR 2026).} We adopt the official pre-LLM implementation of AOT, setting \texttt{VISUAL\_TOKEN\_NUM} (intra-frame token count) to $196/126/84/32$ and \texttt{KEEP\_RATIO} (inter-frame retention parameter) to $60\%/23\%/15\%/45\%$ for target retention ratios of $25\%/10\%/5\%/1\%$, respectively. Since AOT performs intra-frame and inter-frame compression separately, these settings do not enforce exact final token budgets. The resulting average retained token counts are $1614/645/320/71$, corresponding to actual retention ratios of $25.73\%/10.28\%/5.10\%/1.13\%$, respectively. We report AOT results on LLaVA-OneVision at these approximately matched retention levels in Table~\ref{tab:comparison_llava_ov}.

\noindent\textbf{FLoC~\cite{floc} (ICLR 2026).} We independently implement the paper's cosine facility-location objective with exact lazy greedy selection. On LLaVA-OneVision, the 32 sampled frames form one temporal block ($T=32$), from which we retain $B=\operatorname{round}(\gamma N)$ original tokens. Ties are broken by the smallest original index, and the selected tokens are restored to their spatiotemporal order without merging. Retention ratios of $35\%/25\%/15\%/10\%/5\%/1\%$ therefore give $2195/1568/941/627/314/63$ visual tokens. We compute cosine similarities without clipping negative values and evaluate one deterministic run per setting under the LLaVA-OneVision inference protocol shared by the other methods.

\subsection{Computational Cost Estimation}

Following prior work~\cite{fastv, pyramiddrop, fastvid}, we report the theoretical floating-point operation (FLOP) counts of the LLM modules when processing visual/video tokens. LLaVA-OneVision~\cite{llava-onevision} and LLaVA-Video~\cite{llava-video} are built on Qwen2~\cite{qwen2}, whereas Qwen3-VL~\cite{qwen3-vl} is based on Qwen3~\cite{qwen3}. Both decoder families use grouped-query attention~\cite{gqa} and a SwiGLU feed-forward network~\cite{glu}.

For these decoder architectures, the per-layer FLOP count for visual tokens is
\begin{equation}
4nD(h_{kv}d) + 4nD^2 + 4n^2D + 6nDD',
\end{equation}
where $n$ is the number of visual/video tokens, $D$ is the hidden dimension, $D'$ is the FFN intermediate dimension, $h_{kv}$ is the number of key/value heads, and $d$ is the per-head dimension. Each multiplication or addition counts as one FLOP, and a fused multiply--add counts as two. Table~\ref{tab:efficiency} reports these matrix-operation counts in TFLOPs ($10^{12}$ FLOPs), with the KDE counting convention in Table~\ref{tab:ann_kde}.

\subsection{Latency Measurement Protocol and Runtime Details}
\label{appendix:latency_protocol}

We measure the latencies in Tables~\ref{tab:efficiency} and~\ref{tab:efficiency_qwen3_vl} using LMMs-Eval at batch size 1 on a single NVIDIA A100-80GB GPU. For each backbone, all methods process the same 100-sample VideoMME subset with identical prompt and generation configurations, and we report mean latency. Compression time covers the complete pre-LLM stage, including token scoring, clustering, and index remapping where required by each implementation. The reported total prefill latency combines this compression time with LLM prefill time.

The timing scope isolates compression and prefill, the stages directly affected by visual-token compression, and excludes video decoding, vision encoding, and autoregressive output decoding. Video decoding and vision encoding are shared across methods under our matched-input protocol, with runtime also depending on frame batching, parallel processing, and scheduling. The reported speedups therefore describe compression plus prefill rather than full video inference.

For the measurements in Tables~\ref{tab:efficiency} and~\ref{tab:efficiency_qwen3_vl}, we construct content-adaptive intervals from adjacent-frame feature differences, then perform exact density estimation, pivotal sampling, and interval-local weighted merging. Density estimation uses standard PyTorch operations on the full pairwise Gaussian-kernel matrix. The same implementation is used for the 16K/32K-token Qwen3-VL measurements.

\section{Detailed Experimental Results\label{appendix:detailed_results}}

Tables~\ref{tab:comparison_llava_ov_full}--\ref{tab:comparison_internvl} provide benchmark-level results for four Video LLM backbones: LLaVA-OneVision~\cite{llava-onevision}, LLaVA-Video~\cite{llava-video}, Qwen3-VL~\cite{qwen3-vl}, and InternVL3.5~\cite{internvl3.5}. Unless otherwise stated, all main comparisons use the same hyperparameter setting across benchmarks and models.

The first three backbones are evaluated at six retention ratios, from $35\%$ to $1\%$, while InternVL3.5 is evaluated at $10\%$ and $1\%$. These comparisons examine performance as the token budget shrinks and transfer across pretrained models with different visual encoders and tokenization schemes.

\noindent\textbf{Accuracy and aggregation.}
The absolute mean is the unweighted average of MVBench, LongVideoBench, MLVU, and VideoMME overall accuracy. We additionally report VideoMME's short-, medium-, and long-video subsets to show how performance varies with video length, while counting VideoMME only once in the mean. Relative accuracy expresses this mean as a percentage of the corresponding uncompressed mean, making it easier to compare how much performance compression preserves across backbones with different starting accuracies.

\noindent\textbf{Retention ratios.}
A single retention ratio denotes the fraction of original visual tokens passed to the LLM after compression. For methods that also prune inside the LLM, the paired notation $A\%/B\%$ gives the pre-LLM and subsequent retention ratios, both relative to the original token count. Each retention group includes Random as a reference for uniform token selection, allowing methods to be compared against the same simple allocation rule at each budget. Appendix~\ref{appendix:seed_stability} examines variability across random sampling seeds.

\begin{table*}[tp]
\centering
\caption{\textbf{Performance comparison across video benchmarks on \emph{LLaVA-OneVision}.} Best results under each retention ratio are highlighted in bold. The ``A\%/B\%'' retention ratio denotes that A\% of the original visual tokens are preserved, and then further compressed to B\% during the LLM forward pass. This table reports benchmark-level results across six retention ratios, including 35\% and 15\%. Stochastic methods are averaged over 10 random seeds. Parentheses show differences from Random at the same retention ratio, in percentage points (pp) of relative accuracy. Accuracies are percentages, and relative accuracy is the four-benchmark mean normalized by the uncompressed mean.}
\label{tab:comparison_llava_ov_full}
\papertablestyle
\begin{adjustbox}{max width=\linewidth}
\begin{tabular}{@{}l*{10}{c}@{}}
\toprule
\multirow{2}{*}{Method} & \multirow{2}{*}{\shortstack{Retention\\Ratio $\gamma$}} & \multirow{2}{*}{MVBench} & \multirow{2}{*}{LongVB} & \multirow{2}{*}{MLVU} & \multicolumn{4}{c}{VideoMME} & \multicolumn{2}{c@{}}{Mean accuracy} \\
\cmidrule(lr){6-9}\cmidrule(lr){10-11}
 &  &  &  &  & Overall & Short & Medium & Long & Absolute & \multicolumn{1}{c@{}}{Relative} \\
Duration
&  
& 16s
& 1$\sim$60min
& 3$\sim$120min
& 1$\sim$60min
& 1$\sim$3min
& 3$\sim$30min
& 30$\sim$60min
&  
&  
\\
\midrule
\textbf{LLaVA-OV-7B} & 100\% & 58.3 & 56.6 & 63.1 & 58.4 & 70.3 & 56.3 & 48.6 & 59.1 & 100.0\% \\
\midrule
\rowcolor{gray!6}
Random & 35\% & 56.4 & 57.0 & 62.3 & 57.8 & 70.2 & 55.3 & 47.8 & 58.4 & 98.8\% \\
FastV~\cite{fastv} & 100\%/35\% & 57.5 & 56.4 & 61.6 & 57.5 & 70.2 & 54.9 & 47.4 & 58.2 & 98.5\%~(\textcolor{red}{-0.3~pp}) \\
DyCoke~\cite{dycoke} & 35\% & 58.2 & 56.5 & 61.2 & 57.8 & 69.6 & 55.8 & 48.1 & 58.4 & 98.8\%~(0.0~pp) \\
VidCom2~\cite{vidcom2} & 35\% & 56.8 & 57.4 & \textbf{63.1} & 58.3 & 70.6 & 56.3 & 47.9 & 58.9 & 99.7\%~(\textcolor{green!50!black}{+0.9~pp}) \\
PruneVID~\cite{prunevid} & 35\% & 57.4 & 56.0 & 62.6 & 57.6 & 69.0 & 55.4 & 48.4 & 58.4 & 98.8\%~(0.0~pp) \\
VisionZip~\cite{visionzip} & 35\% & 58.2 & 56.3 & 62.9 & 58.2 & \textbf{70.8} & 54.8 & \textbf{48.9} & 58.9 & 99.7\%~(\textcolor{green!50!black}{+0.9~pp}) \\
FastVID~\cite{fastvid} & 35\% & \textbf{58.4} & 55.9 & 61.7 & \textbf{58.4} & \textbf{70.8} & \textbf{56.4} & 48.0 & 58.6 & 99.2\%~(\textcolor{green!50!black}{+0.4~pp}) \\
FLoC~\cite{floc} & 35\% & 57.7 & 56.5 & 62.7 & 57.1 & 69.3 & 54.3 & 47.8 & 58.5 & 99.0\%~(\textcolor{green!50!black}{+0.2~pp}) \\
\textbf{KiToke (Ours)} & 35\% & 58.2 & \textbf{57.5} & 62.9 & 58.0 & 70.4 & 54.9 & 48.6 & \textbf{59.1} & \textbf{100.0\%}~\textbf{(\textcolor{green!50!black}{+1.2~pp})} \\
\midrule
\rowcolor{gray!6}
Random & 25\% & 56.7 & 57.1 & 61.9 & 56.7 & 68.6 & 54.4 & 47.2 & 58.1 & 98.3\% \\
FastV~\cite{fastv} & 100\%/25\% & 57.2 & 55.0 & 60.8 & 57.3 & 70.0 & 54.0 & 47.9 & 57.6 & 97.5\%~(\textcolor{red}{-0.8~pp}) \\
DyCoke~\cite{dycoke} & 25\% & 57.4 & 53.8 & 58.4 & 53.9 & 64.1 & 52.7 & 44.8 & 55.9 & 94.6\%~(\textcolor{red}{-3.7~pp}) \\
VidCom2~\cite{vidcom2} & 25\% & 57.0 & 56.3 & 62.4 & 57.4 & 68.1 & 56.0 & 48.2 & 58.3 & 98.6\%~(\textcolor{green!50!black}{+0.3~pp}) \\
PruneVID~\cite{prunevid} & 25\% & 56.8 & 55.6 & 62.8 & 56.9 & 68.7 & 54.1 & 48.0 & 58.0 & 98.1\%~(\textcolor{red}{-0.2~pp}) \\
VisionZip~\cite{visionzip} & 25\% & 57.7 & 56.0 & 62.1 & 57.7 & 69.6 & 56.0 & 47.5 & 58.4 & 98.8\%~(\textcolor{green!50!black}{+0.5~pp}) \\
FastVID~\cite{fastvid} & 25\% & 58.0 & 56.0 & 61.7 & 58.2 & \textbf{70.3} & \textbf{56.4} & 47.9 & 58.5 & 99.0\%~(\textcolor{green!50!black}{+0.7~pp}) \\
FLoC~\cite{floc} & 25\% & 56.7 & 56.8 & 62.1 & 57.0 & 68.8 & 54.9 & 47.3 & 58.2 & 98.5\%~(\textcolor{green!50!black}{+0.2~pp}) \\
\textbf{KiToke (Ours)} & 25\% & \textbf{58.4} & \textbf{57.4} & \textbf{63.4} & \textbf{58.3} & 70.0 & 56.1 & \textbf{48.8} & \textbf{59.4} & \textbf{100.5\%}~\textbf{(\textcolor{green!50!black}{+2.2~pp})} \\
\midrule
\rowcolor{gray!6}
Random & 15\% & 56.0 & 56.1 & 61.1 & 56.1 & 67.2 & 53.3 & 47.7 & 57.3 & 97.0\% \\
FastV~\cite{fastv} & 100\%/15\% & 55.5 & 53.0 & 59.7 & 55.3 & 67.0 & 52.8 & 46.0 & 55.9 & 94.6\%~(\textcolor{red}{-2.4~pp}) \\
VidCom2~\cite{vidcom2} & 15\% & 54.3 & 53.0 & 60.9 & 56.3 & 65.4 & 55.2 & 48.1 & 56.1 & 94.9\%~(\textcolor{red}{-2.1~pp}) \\
PruneVID~\cite{prunevid} & 15\% & 56.1 & 55.5 & 62.4 & 56.6 & 67.8 & 53.9 & 48.2 & 57.7 & 97.6\%~(\textcolor{green!50!black}{+0.6~pp}) \\
VisionZip~\cite{visionzip} & 15\% & 57.2 & 54.5 & 61.0 & 56.9 & 67.8 & 55.7 & 47.3 & 57.4 & 97.1\%~(\textcolor{green!50!black}{+0.1~pp}) \\
FastVID~\cite{fastvid} & 15\% & 57.3 & 55.8 & 60.5 & 57.4 & \textbf{69.3} & 55.9 & 46.9 & 57.8 & 97.8\%~(\textcolor{green!50!black}{+0.8~pp}) \\
FLoC~\cite{floc} & 15\% & 55.8 & 55.2 & 61.3 & 56.4 & 67.3 & 54.3 & 47.7 & 57.2 & 96.8\%~(\textcolor{red}{-0.2~pp}) \\
\textbf{KiToke (Ours)} & 15\% & \textbf{57.6} & \textbf{56.5} & \textbf{62.5} & \textbf{57.7} & 68.8 & \textbf{56.0} & \textbf{48.3} & \textbf{58.6} & \textbf{99.2\%}~\textbf{(\textcolor{green!50!black}{+2.2~pp})} \\
\midrule
\rowcolor{gray!6}
Random & 10\% & 54.2 & 54.7 & 59.7 & 54.7 & 64.8 & 53.3 & 45.9 & 55.8 & 94.4\% \\
FastV~\cite{fastv} & 100\%/10\% & 53.0 & 50.7 & 57.1 & 53.6 & 63.0 & 51.9 & 46.0 & 53.6 & 90.7\%~(\textcolor{red}{-3.7~pp}) \\
VidCom2~\cite{vidcom2} & 10\% & 52.5 & 50.5 & 58.0 & 53.3 & 61.8 & 51.4 & 46.8 & 53.6 & 90.7\%~(\textcolor{red}{-3.7~pp}) \\
PruneVID~\cite{prunevid} & 10\% & 55.7 & 54.8 & 60.7 & 55.9 & \textbf{66.8} & 54.0 & 47.0 & 56.8 & 96.1\%~(\textcolor{green!50!black}{+1.7~pp}) \\
VisionZip~\cite{visionzip} & 10\% & 55.6 & 52.7 & 59.4 & 55.6 & 66.3 & 54.2 & 46.4 & 55.8 & 94.4\%~(0.0~pp) \\
FastVID~\cite{fastvid} & 10\% & 56.6 & 55.6 & 60.4 & 55.6 & 66.2 & 54.2 & 46.4 & 57.0 & 96.4\%~(\textcolor{green!50!black}{+2.0~pp}) \\
FLoC~\cite{floc} & 10\% & 55.6 & 54.8 & 60.6 & 55.4 & 66.1 & 53.7 & 46.4 & 56.6 & 95.8\%~(\textcolor{green!50!black}{+1.4~pp}) \\
\textbf{KiToke (Ours)} & 10\% & \textbf{57.2} & \textbf{57.4} & \textbf{62.1} & \textbf{56.0} & 66.3 & \textbf{54.4} & \textbf{47.1} & \textbf{58.2} & \textbf{98.5\%}~\textbf{(\textcolor{green!50!black}{+4.1~pp})} \\
\midrule
\rowcolor{gray!6}
Random & 5\% & 53.0 & 53.9 & 58.2 & 52.5 & 60.2 & 51.1 & 46.1 & 54.4 & 92.0\% \\
FastV~\cite{fastv} & 100\%/5\% & 49.9 & 48.3 & 55.4 & 51.5 & 58.0 & 50.9 & 45.7 & 51.3 & 86.8\%~(\textcolor{red}{-5.2~pp}) \\
VidCom2~\cite{vidcom2} & 5\% & 46.9 & 47.4 & 53.6 & 49.4 & 55.7 & 49.4 & 43.1 & 49.3 & 83.4\%~(\textcolor{red}{-8.6~pp}) \\
PruneVID~\cite{prunevid} & 5\% & 53.3 & 52.9 & 58.7 & 53.0 & 62.3 & 50.6 & 46.0 & 54.5 & 92.2\%~(\textcolor{green!50!black}{+0.2~pp}) \\
VisionZip~\cite{visionzip} & 5\% & 51.9 & 48.6 & 56.6 & 52.4 & 59.9 & 51.6 & 45.9 & 52.4 & 88.7\%~(\textcolor{red}{-3.3~pp}) \\
FastVID~\cite{fastvid} & 5\% & 54.8 & 53.7 & 58.3 & 52.6 & 62.6 & 51.0 & 44.3 & 54.8 & 92.7\%~(\textcolor{green!50!black}{+0.7~pp}) \\
FLoC~\cite{floc} & 5\% & 53.5 & 52.3 & 57.5 & 52.8 & 62.4 & 50.4 & 45.4 & 54.0 & 91.4\%~(\textcolor{red}{-0.6~pp}) \\
\textbf{KiToke (Ours)} & 5\% & \textbf{55.2} & \textbf{54.9} & \textbf{60.8} & \textbf{54.1} & \textbf{63.0} & \textbf{52.2} & \textbf{47.1} & \textbf{56.2} & \textbf{95.1\%}~\textbf{(\textcolor{green!50!black}{+3.1~pp})} \\
\midrule
\rowcolor{gray!6}
Random & 1\% & 48.0 & 46.9 & 54.6 & 49.0 & 54.4 & 49.1 & 43.3 & 49.6 & 83.9\% \\
FastV~\cite{fastv} & 100\%/1\% & 43.8 & 46.7 & 51.7 & 48.0 & 52.7 & 48.8 & 42.7 & 47.6 & 80.5\%~(\textcolor{red}{-3.4~pp}) \\
VidCom2~\cite{vidcom2} & 1\% & 39.2 & 41.7 & 45.7 & 41.7 & 42.0 & 42.1 & 40.9 & 42.1 & 71.2\%~(\textcolor{red}{-12.7~pp}) \\
PruneVID~\cite{prunevid} & 1\% & 46.0 & 46.7 & 53.7 & 48.0 & 53.0 & 47.7 & 43.3 & 48.6 & 82.2\%~(\textcolor{red}{-1.7~pp}) \\
VisionZip~\cite{visionzip} & 1\% & 42.6 & 43.9 & 52.1 & 46.5 & 49.6 & 47.3 & 42.7 & 46.3 & 78.3\%~(\textcolor{red}{-5.6~pp}) \\
FastVID~\cite{fastvid} & 1\% & 45.7 & 46.1 & 51.9 & 46.9 & 52.3 & 46.4 & 41.8 & 47.6 & 80.5\%~(\textcolor{red}{-3.4~pp}) \\
FLoC~\cite{floc} & 1\% & 47.3 & 47.0 & 51.4 & 47.7 & 54.8 & 47.0 & 41.3 & 48.4 & 81.9\%~(\textcolor{red}{-2.0~pp}) \\
\textbf{KiToke (Ours)} & 1\% & \textbf{52.5} & \textbf{51.1} & \textbf{56.9} & \textbf{50.2} & \textbf{57.4} & \textbf{49.3} & \textbf{43.8} & \textbf{52.7} & \textbf{89.2\%}~\textbf{(\textcolor{green!50!black}{+5.3~pp})} \\
\bottomrule
\end{tabular}
\end{adjustbox}
\appendixanalysis{\textbf{LLaVA-OneVision.} LLaVA-OneVision preserves the largest fraction of its uncompressed accuracy among the evaluated backbones, and KiToke achieves the highest mean accuracy at every tested retention ratio in Table~\ref{tab:comparison_llava_ov_full}. At 15\% retention, it preserves 99.2\% of uncompressed accuracy, compared with 97.8\% for the strongest prior method and 97.0\% for Random. Its advantage over Random grows from 2.2 points at 15\% retention to 5.3 points at 1\%, where all compared prior methods fall below Random. Appendix~\ref{appendix:selection_fixed_merging} examines this low-budget setting through controlled comparisons of the allocation rule.}
\end{table*}

\begin{table*}[tp]
\centering
\caption{\textbf{Performance comparison across video benchmarks on \emph{LLaVA-Video}.} Best results under each retention ratio are highlighted in bold. Stochastic methods are averaged over 10 random seeds. Parentheses show differences from Random at the same retention ratio, in percentage points (pp) of relative accuracy. Accuracies are percentages, and relative accuracy is the four-benchmark mean normalized by the uncompressed mean.}
\label{tab:comparison_llava_video}
\papertablestyle
\begin{adjustbox}{max width=\linewidth}
\begin{tabular}{@{}l*{10}{c}@{}}
\toprule
\multirow{2}{*}{Method} & \multirow{2}{*}{\shortstack{Retention\\Ratio $\gamma$}} & \multirow{2}{*}{MVBench} & \multirow{2}{*}{LongVB} & \multirow{2}{*}{MLVU} & \multicolumn{4}{c}{VideoMME} & \multicolumn{2}{c@{}}{Mean accuracy} \\
\cmidrule(lr){6-9}\cmidrule(lr){10-11}
 &  &  &  &  & Overall & Short & Medium & Long & Absolute & \multicolumn{1}{c@{}}{Relative} \\
Duration
&  
& 16s
& 1$\sim$60min
& 3$\sim$120min
& 1$\sim$60min
& 1$\sim$3min
& 3$\sim$30min
& 30$\sim$60min
&  
&  
\\
\midrule
\textbf{LLaVA-Video-7B} & 100\% & 60.4 & 58.7 & 67.4 & 64.2 & 77.2 & 62.2 & 53.1 & 62.7 & 100.0\% \\
\midrule
\rowcolor{gray!6}
Random & 35\% & 57.7 & 56.2 & 63.8 & 61.1 & 73.7 & 60.3 & 49.0 & 59.7 & 95.2\% \\
FastV~\cite{fastv} & 100\%/35\% & 59.3 & 57.5 & 66.3 & 62.8 & 75.5 & \textbf{62.0} & 51.0 & 61.5 & 98.1\%~(\textcolor{green!50!black}{+2.9~pp}) \\
DyCoke~\cite{dycoke} & 35\% & 58.3 & 56.5 & 60.4 & 62.4 & 75.6 & 60.6 & 51.1 & 59.4 & 94.7\%~(\textcolor{red}{-0.5~pp}) \\
VidCom2~\cite{vidcom2} & 35\% & 57.9 & 58.1 & 61.9 & \textbf{63.0} & \textbf{76.4} & 61.4 & 51.0 & 60.2 & 96.0\%~(\textcolor{green!50!black}{+0.8~pp}) \\
PruneVID~\cite{prunevid} & 35\% & 56.3 & 57.9 & 65.7 & 61.3 & 74.4 & 59.2 & 50.3 & 60.3 & 96.2\%~(\textcolor{green!50!black}{+1.0~pp}) \\
VisionZip~\cite{visionzip} & 35\% & 56.8 & 57.4 & 66.1 & 61.8 & 74.7 & 60.0 & 50.7 & 60.5 & 96.5\%~(\textcolor{green!50!black}{+1.3~pp}) \\
FastVID~\cite{fastvid} & 35\% & 59.2 & 57.2 & 66.5 & 62.3 & 75.2 & 61.2 & 50.5 & 61.3 & 97.8\%~(\textcolor{green!50!black}{+2.6~pp}) \\
\textbf{KiToke (Ours)} & 35\% & \textbf{59.5} & \textbf{58.3} & \textbf{67.3} & \textbf{63.0} & 75.8 & 61.8 & \textbf{51.3} & \textbf{62.0} & \textbf{98.9\%}~\textbf{(\textcolor{green!50!black}{+3.7~pp})} \\
\midrule
\rowcolor{gray!6}
Random & 25\% & 56.9 & 55.9 & 61.5 & 60.1 & 73.1 & 58.2 & 48.8 & 58.6 & 93.5\% \\
FastV~\cite{fastv} & 100\%/25\% & 57.6 & 57.2 & 65.8 & 61.8 & 74.5 & 60.5 & 50.5 & 60.6 & 96.7\%~(\textcolor{green!50!black}{+3.2~pp}) \\
DyCoke~\cite{dycoke} & 25\% & 56.6 & 56.2 & 56.8 & 59.4 & 72.3 & 56.0 & 49.8 & 57.2 & 91.2\%~(\textcolor{red}{-2.3~pp}) \\
VidCom2~\cite{vidcom2} & 25\% & 57.0 & 56.5 & 58.6 & 61.4 & 73.2 & 60.7 & 50.2 & 58.4 & 93.1\%~(\textcolor{red}{-0.4~pp}) \\
PruneVID~\cite{prunevid} & 25\% & 56.2 & 57.7 & 63.7 & 59.9 & 72.4 & 56.9 & 50.3 & 59.4 & 94.7\%~(\textcolor{green!50!black}{+1.2~pp}) \\
VisionZip~\cite{visionzip} & 25\% & 56.4 & 56.4 & 64.2 & 61.4 & 74.5 & 59.4 & 50.4 & 59.6 & 95.1\%~(\textcolor{green!50!black}{+1.6~pp}) \\
FastVID~\cite{fastvid} & 25\% & 58.1 & 57.1 & 65.3 & 61.8 & 74.3 & 60.7 & 50.5 & 60.6 & 96.7\%~(\textcolor{green!50!black}{+3.2~pp}) \\
\textbf{KiToke (Ours)} & 25\% & \textbf{58.7} & \textbf{58.2} & \textbf{66.5} & \textbf{62.1} & \textbf{74.6} & \textbf{60.8} & \textbf{51.0} & \textbf{61.4} & \textbf{97.9\%}~\textbf{(\textcolor{green!50!black}{+4.4~pp})} \\
\midrule
\rowcolor{gray!6}
Random & 15\% & 55.3 & 55.9 & 59.6 & 58.7 & 70.6 & 56.2 & 49.4 & 57.4 & 91.5\% \\
FastV~\cite{fastv} & 100\%/15\% & 53.5 & 55.7 & 63.1 & 60.2 & 71.4 & 59.4 & 49.7 & 58.1 & 92.7\%~(\textcolor{green!50!black}{+1.2~pp}) \\
VidCom2~\cite{vidcom2} & 15\% & 53.4 & 52.4 & 56.7 & 58.3 & 68.1 & 56.9 & 49.8 & 55.2 & 88.0\%~(\textcolor{red}{-3.5~pp}) \\
PruneVID~\cite{prunevid} & 15\% & 54.9 & 56.3 & 61.7 & 58.6 & 70.2 & 56.4 & 49.2 & 57.9 & 92.3\%~(\textcolor{green!50!black}{+0.8~pp}) \\
VisionZip~\cite{visionzip} & 15\% & 55.6 & 55.0 & 62.0 & 60.1 & 71.0 & 59.6 & 49.7 & 58.2 & 92.8\%~(\textcolor{green!50!black}{+1.3~pp}) \\
FastVID~\cite{fastvid} & 15\% & 57.6 & 55.7 & 64.6 & 60.6 & 72.3 & 59.7 & 49.8 & 59.6 & 95.1\%~(\textcolor{green!50!black}{+3.6~pp}) \\
\textbf{KiToke (Ours)} & 15\% & \textbf{58.2} & \textbf{58.0} & \textbf{66.1} & \textbf{61.0} & \textbf{72.9} & \textbf{60.1} & \textbf{50.0} & \textbf{60.8} & \textbf{97.0\%}~\textbf{(\textcolor{green!50!black}{+5.5~pp})} \\
\midrule
\rowcolor{gray!6}
Random & 10\% & 54.1 & 55.1 & 59.2 & 57.8 & 69.0 & 56.8 & 47.6 & 56.6 & 90.3\% \\
FastV~\cite{fastv} & 100\%/10\% & 51.4 & 53.1 & 62.8 & 58.2 & 68.4 & 57.2 & 49.0 & 56.4 & 90.0\%~(\textcolor{red}{-0.3~pp}) \\
VidCom2~\cite{vidcom2} & 10\% & 50.8 & 50.7 & 54.8 & 56.9 & 66.6 & 57.0 & 47.2 & 53.3 & 85.0\%~(\textcolor{red}{-5.3~pp}) \\
PruneVID~\cite{prunevid} & 10\% & 54.1 & 55.2 & 61.0 & 57.3 & 68.9 & 54.7 & 48.2 & 56.9 & 90.7\%~(\textcolor{green!50!black}{+0.4~pp}) \\
VisionZip~\cite{visionzip} & 10\% & 55.0 & 52.1 & 60.3 & 57.0 & 67.7 & 55.4 & 47.8 & 56.1 & 89.5\%~(\textcolor{red}{-0.8~pp}) \\
FastVID~\cite{fastvid} & 10\% & 56.5 & 55.1 & 63.0 & 58.0 & 69.0 & 56.7 & 48.3 & 58.2 & 92.8\%~(\textcolor{green!50!black}{+2.5~pp}) \\
\textbf{KiToke (Ours)} & 10\% & \textbf{57.2} & \textbf{57.0} & \textbf{65.1} & \textbf{59.1} & \textbf{70.7} & \textbf{57.7} & \textbf{49.1} & \textbf{59.6} & \textbf{95.1\%}~\textbf{(\textcolor{green!50!black}{+4.8~pp})} \\
\midrule
\rowcolor{gray!6}
Random & 5\% & 52.9 & 52.7 & 57.3 & 55.4 & 67.2 & 52.7 & 46.2 & 54.6 & 87.1\% \\
FastV~\cite{fastv} & 100\%/5\% & 49.0 & 50.3 & 57.7 & 55.7 & 66.4 & 53.4 & 47.3 & 53.2 & 84.8\%~(\textcolor{red}{-2.3~pp}) \\
VidCom2~\cite{vidcom2} & 5\% & 46.5 & 47.0 & 52.5 & 52.8 & 57.9 & 53.4 & 47.1 & 49.7 & 79.3\%~(\textcolor{red}{-7.8~pp}) \\
PruneVID~\cite{prunevid} & 5\% & 52.7 & 52.5 & 58.3 & 55.3 & 66.3 & 52.2 & 47.4 & 54.7 & 87.2\%~(\textcolor{green!50!black}{+0.1~pp}) \\
VisionZip~\cite{visionzip} & 5\% & 53.1 & 48.1 & 51.8 & 56.7 & 67.9 & 54.8 & 47.4 & 52.4 & 83.6\%~(\textcolor{red}{-3.5~pp}) \\
FastVID~\cite{fastvid} & 5\% & 55.0 & 53.8 & 58.0 & 56.0 & 67.3 & 54.3 & 46.4 & 55.7 & 88.8\%~(\textcolor{green!50!black}{+1.7~pp}) \\
\textbf{KiToke (Ours)} & 5\% & \textbf{55.6} & \textbf{55.7} & \textbf{62.5} & \textbf{57.4} & \textbf{69.3} & \textbf{54.9} & \textbf{48.0} & \textbf{57.8} & \textbf{92.2\%}~\textbf{(\textcolor{green!50!black}{+5.1~pp})} \\
\midrule
\rowcolor{gray!6}
Random & 1\% & 47.6 & 48.9 & 53.1 & 49.5 & 56.2 & 49.1 & 43.2 & 49.8 & 79.4\% \\
FastV~\cite{fastv} & 100\%/1\% & 46.7 & 46.1 & 51.3 & 51.0 & 57.0 & 51.1 & 45.0 & 48.8 & 77.8\%~(\textcolor{red}{-1.6~pp}) \\
VidCom2~\cite{vidcom2} & 1\% & 41.9 & 44.4 & 49.1 & 45.6 & 46.9 & 47.4 & 42.4 & 45.2 & 72.1\%~(\textcolor{red}{-7.3~pp}) \\
PruneVID~\cite{prunevid} & 1\% & 44.0 & 47.4 & 52.8 & 49.6 & 53.8 & 49.7 & 45.2 & 48.5 & 77.4\%~(\textcolor{red}{-2.0~pp}) \\
VisionZip~\cite{visionzip} & 1\% & 43.3 & 45.9 & 50.6 & 49.7 & 53.4 & 51.0 & 44.6 & 47.4 & 75.6\%~(\textcolor{red}{-3.8~pp}) \\
FastVID~\cite{fastvid} & 1\% & 45.0 & 48.2 & 52.0 & 51.1 & 58.0 & 50.3 & 45.0 & 49.1 & 78.3\%~(\textcolor{red}{-1.1~pp}) \\
\textbf{KiToke (Ours)} & 1\% & \textbf{49.0} & \textbf{51.9} & \textbf{55.8} & \textbf{52.3} & \textbf{60.0} & \textbf{51.3} & \textbf{45.7} & \textbf{52.3} & \textbf{83.4\%}~\textbf{(\textcolor{green!50!black}{+4.0~pp})} \\
\bottomrule
\end{tabular}
\end{adjustbox}
\appendixanalysis{\textbf{LLaVA-Video.} KiToke also achieves the highest mean accuracy at every tested retention ratio on LLaVA-Video, as shown in Table~\ref{tab:comparison_llava_video}. This backbone loses more relative accuracy than LLaVA-OneVision at low retention and uses newline tokens after each row of 13 spatial tokens, requiring compression to retain this layout structure alongside the visual content. As the budget shrinks, prior heuristics that perform well at moderate retention often fall below Random. KiToke maintains its advantage, exceeding Random by $+5.5$ and $+4.0$ points at 15\% and 1\% retention, respectively.}
\end{table*}

\begin{table*}[tp]
\centering
\caption{\textbf{Performance comparison across video benchmarks on \emph{Qwen3-VL}.} Best results under each retention ratio are highlighted in bold. Stochastic methods are averaged over 10 random seeds. Parentheses show differences from Random at the same retention ratio, in percentage points (pp) of relative accuracy. Accuracies are percentages, and relative accuracy is the four-benchmark mean normalized by the uncompressed mean.}
\label{tab:comparison_qwen3_vl}
\papertablestyle
\begin{adjustbox}{max width=\linewidth}
\begin{tabular}{@{}l*{10}{c}@{}}
\toprule
\multirow{2}{*}{Method} & \multirow{2}{*}{\shortstack{Retention\\Ratio $\gamma$}} & \multirow{2}{*}{MVBench} & \multirow{2}{*}{LongVB} & \multirow{2}{*}{MLVU} & \multicolumn{4}{c}{VideoMME} & \multicolumn{2}{c@{}}{Mean accuracy} \\
\cmidrule(lr){6-9}\cmidrule(lr){10-11}
 &  &  &  &  & Overall & Short & Medium & Long & Absolute & \multicolumn{1}{c@{}}{Relative} \\
Duration
&  
& 16s
& 1$\sim$60min
& 3$\sim$120min
& 1$\sim$60min
& 1$\sim$3min
& 3$\sim$30min
& 30$\sim$60min
&  
&  
\\
\midrule
\textbf{Qwen3-VL-8B} & 100\% & 68.5 & 62.5 & 71.4 & 68.7 & 79.9 & 67.0 & 59.1 & 67.8 & 100.0\% \\
\midrule
\rowcolor{gray!6}
Random & 35\% & 66.6 & 61.0 & 69.3 & 66.3 & 77.4 & 64.9 & 56.4 & 65.8 & 97.1\% \\
DyCoke~\cite{dycoke} & 35\% & 64.1 & 60.1 & 67.5 & \textbf{67.0} & 79.0 & \textbf{65.4} & 56.4 & 64.7 & 95.4\%~(\textcolor{red}{-1.7~pp}) \\
VidCom2~\cite{vidcom2} & 35\% & 66.3 & 61.1 & 68.7 & 66.9 & \textbf{79.1} & 65.3 & 56.3 & 65.8 & 97.1\%~(0.0~pp) \\
PruneVID~\cite{prunevid} & 35\% & 60.9 & 60.4 & 67.6 & 63.4 & 75.0 & 59.7 & 55.7 & 63.1 & 93.1\%~(\textcolor{red}{-4.0~pp}) \\
VisionZip~\cite{visionzip} & 35\% & 67.0 & 61.2 & 68.8 & 66.3 & 77.8 & 64.0 & 57.0 & 65.8 & 97.1\%~(0.0~pp) \\
FastVID~\cite{fastvid} & 35\% & 66.1 & 60.1 & 68.9 & 64.2 & 74.8 & 63.0 & 54.9 & 64.8 & 95.6\%~(\textcolor{red}{-1.5~pp}) \\
\textbf{KiToke (Ours)} & 35\% & \textbf{67.1} & \textbf{62.4} & \textbf{69.4} & \textbf{67.0} & 78.4 & 65.2 & \textbf{57.2} & \textbf{66.5} & \textbf{98.1\%}~\textbf{(\textcolor{green!50!black}{+1.0~pp})} \\
\midrule
\rowcolor{gray!6}
Random & 25\% & 64.5 & 61.0 & 68.6 & 64.9 & 75.9 & 63.4 & 55.3 & 64.8 & 95.6\% \\
DyCoke~\cite{dycoke} & 25\% & 59.1 & 57.3 & 62.9 & 64.0 & 75.4 & 62.3 & 54.2 & 60.8 & 89.7\%~(\textcolor{red}{-5.9~pp}) \\
VidCom2~\cite{vidcom2} & 25\% & 64.1 & 60.3 & 67.3 & 65.3 & 76.3 & \textbf{64.0} & 55.7 & 64.3 & 94.8\%~(\textcolor{red}{-0.8~pp}) \\
PruneVID~\cite{prunevid} & 25\% & 58.4 & 59.2 & 65.5 & 62.5 & 72.9 & 60.3 & 54.3 & 61.4 & 90.6\%~(\textcolor{red}{-5.0~pp}) \\
VisionZip~\cite{visionzip} & 25\% & \textbf{65.4} & 60.9 & 67.0 & 65.4 & 76.4 & 62.8 & \textbf{56.9} & 64.7 & 95.4\%~(\textcolor{red}{-0.2~pp}) \\
FastVID~\cite{fastvid} & 25\% & 65.1 & 59.7 & 68.2 & 63.5 & 74.3 & 61.4 & 54.8 & 64.1 & 94.5\%~(\textcolor{red}{-1.1~pp}) \\
\textbf{KiToke (Ours)} & 25\% & 65.2 & \textbf{61.5} & \textbf{69.5} & \textbf{65.8} & \textbf{77.6} & 63.9 & 56.0 & \textbf{65.5} & \textbf{96.6\%}~\textbf{(\textcolor{green!50!black}{+1.0~pp})} \\
\midrule
\rowcolor{gray!6}
Random & 15\% & 61.0 & 58.7 & 66.6 & 62.0 & 72.2 & 60.2 & 53.7 & 62.1 & 91.6\% \\
VidCom2~\cite{vidcom2} & 15\% & 58.8 & 57.4 & 64.2 & 62.3 & 72.8 & 60.1 & 53.9 & 60.7 & 89.5\%~(\textcolor{red}{-2.1~pp}) \\
PruneVID~\cite{prunevid} & 15\% & 55.6 & 57.4 & 62.8 & 58.9 & 69.1 & 55.4 & 52.0 & 58.7 & 86.6\%~(\textcolor{red}{-5.0~pp}) \\
VisionZip~\cite{visionzip} & 15\% & 62.8 & 56.0 & 66.0 & 61.6 & 71.9 & 58.9 & 54.1 & 61.6 & 90.9\%~(\textcolor{red}{-0.7~pp}) \\
FastVID~\cite{fastvid} & 15\% & 62.1 & 57.8 & 66.1 & 62.1 & 73.8 & 59.6 & 52.9 & 62.0 & 91.4\%~(\textcolor{red}{-0.2~pp}) \\
\textbf{KiToke (Ours)} & 15\% & \textbf{62.9} & \textbf{60.4} & \textbf{67.6} & \textbf{64.1} & \textbf{75.8} & \textbf{60.8} & \textbf{55.9} & \textbf{63.8} & \textbf{94.1\%}~\textbf{(\textcolor{green!50!black}{+2.5~pp})} \\
\midrule
\rowcolor{gray!6}
Random & 10\% & 57.6 & 57.1 & 65.3 & 61.7 & 72.6 & 59.7 & 52.6 & 60.5 & 89.2\% \\
VidCom2~\cite{vidcom2} & 10\% & 54.5 & 54.7 & 60.3 & 61.6 & 72.3 & 59.4 & 52.9 & 57.8 & 85.3\%~(\textcolor{red}{-3.9~pp}) \\
PruneVID~\cite{prunevid} & 10\% & 53.8 & 55.4 & 60.0 & 56.8 & 65.6 & 53.6 & 51.3 & 56.5 & 83.3\%~(\textcolor{red}{-5.9~pp}) \\
VisionZip~\cite{visionzip} & 10\% & 58.2 & 54.5 & 63.1 & 59.0 & 67.3 & 57.6 & 52.2 & 58.7 & 86.6\%~(\textcolor{red}{-2.6~pp}) \\
FastVID~\cite{fastvid} & 10\% & 59.7 & 56.6 & 64.5 & 61.0 & 71.1 & 58.8 & 53.1 & 60.5 & 89.2\%~(0.0~pp) \\
\textbf{KiToke (Ours)} & 10\% & \textbf{60.5} & \textbf{58.6} & \textbf{67.0} & \textbf{62.8} & \textbf{74.3} & \textbf{60.2} & \textbf{53.9} & \textbf{62.2} & \textbf{91.7\%}~\textbf{(\textcolor{green!50!black}{+2.5~pp})} \\
\midrule
\rowcolor{gray!6}
Random & 5\% & 53.1 & 54.7 & 60.3 & 59.4 & 68.7 & 56.4 & 53.2 & 56.9 & 83.9\% \\
VidCom2~\cite{vidcom2} & 5\% & 49.2 & 50.7 & 55.0 & 56.7 & 63.6 & 54.9 & 51.8 & 52.9 & 78.0\%~(\textcolor{red}{-5.9~pp}) \\
PruneVID~\cite{prunevid} & 5\% & 49.8 & 53.9 & 57.8 & 53.8 & 61.4 & 51.4 & 48.4 & 53.8 & 79.4\%~(\textcolor{red}{-4.5~pp}) \\
VisionZip~\cite{visionzip} & 5\% & 51.7 & 49.6 & 57.2 & 55.5 & 60.7 & 53.8 & 52.1 & 53.5 & 78.9\%~(\textcolor{red}{-5.0~pp}) \\
FastVID~\cite{fastvid} & 5\% & 54.3 & 53.4 & 60.9 & 57.7 & 65.0 & 56.6 & 51.6 & 56.6 & 83.5\%~(\textcolor{red}{-0.4~pp}) \\
\textbf{KiToke (Ours)} & 5\% & \textbf{57.1} & \textbf{56.7} & \textbf{63.7} & \textbf{59.9} & \textbf{69.8} & \textbf{56.7} & \textbf{53.3} & \textbf{59.3} & \textbf{87.5\%}~\textbf{(\textcolor{green!50!black}{+3.6~pp})} \\
\midrule
\rowcolor{gray!6}
Random & 1\% & 45.9 & 48.2 & 53.8 & 52.7 & 58.9 & 50.1 & 49.1 & 50.2 & 74.0\% \\
VidCom2~\cite{vidcom2} & 1\% & 41.7 & 45.3 & 47.2 & 46.4 & 45.3 & 46.8 & 47.1 & 45.1 & 66.5\%~(\textcolor{red}{-7.5~pp}) \\
PruneVID~\cite{prunevid} & 1\% & 42.6 & 48.1 & 52.1 & 48.4 & 52.1 & 47.2 & 45.8 & 47.8 & 70.5\%~(\textcolor{red}{-3.5~pp}) \\
VisionZip~\cite{visionzip} & 1\% & 41.4 & 45.8 & 49.4 & 47.1 & 48.2 & 46.2 & 46.8 & 45.9 & 67.7\%~(\textcolor{red}{-6.3~pp}) \\
FastVID~\cite{fastvid} & 1\% & 44.0 & 46.5 & 54.0 & 49.5 & 53.8 & 48.8 & 46.0 & 48.5 & 71.5\%~(\textcolor{red}{-2.5~pp}) \\
\textbf{KiToke (Ours)} & 1\% & \textbf{48.2} & \textbf{50.7} & \textbf{56.3} & \textbf{53.6} & \textbf{60.9} & \textbf{50.2} & \textbf{49.7} & \textbf{52.2} & \textbf{77.0\%}~\textbf{(\textcolor{green!50!black}{+3.0~pp})} \\
\bottomrule
\end{tabular}
\end{adjustbox}
\appendixanalysis{\textbf{Qwen3-VL.} The Qwen3-VL evaluation in Table~\ref{tab:comparison_qwen3_vl} tests transfer to a different architecture and tokenization pipeline. Qwen3-VL uses timestamp encoding with Interleaved-MRoPE and retains a smaller fraction of uncompressed accuracy than LLaVA-OneVision at low retention. Prior methods fall below Random at almost every retention ratio, showing that their gains on the LLaVA backbones do not consistently transfer to Qwen3-VL. Without backbone-specific tuning, KiToke leads in mean accuracy, exceeding Random by $+2.5$ points at 15\% retention and $+3.0$ at 1\%.}
\end{table*}

\subsection{Open-ended and Cross-domain QA}\label{appendix:open_ended_cross_domain}

\noindent\textbf{Metrics and evaluated data.}
All four benchmarks are evaluated on their test splits. EgoSchema uses the 500-question public subset and measures option-selection accuracy. Video-MMMU contains 900 questions, divided equally among perception, comprehension, and adaptation. We average accuracy over all questions using the benchmark's option matching and normalized text/numerical answer matching. For open-ended QA, CVRR-ES contains 2,400 questions across 11 reasoning dimensions and reports the unweighted mean of their judged correctness rates, while VideoEval-Pro reports the fraction of its 1,289 questions judged correct. We express all accuracies as percentages.

\noindent\textbf{Answer generation.}
We use LLaVA-OneVision-7B with 32 uniformly sampled frames, the \texttt{qwen\_1\_5} chat template, and batch size 1. Each question receives one answer, generated with at most 256 new tokens, temperature 0, top-$p=1$, one beam, and sampling disabled. The task configurations use a double-newline stop delimiter.

\noindent\textbf{Judging protocol.}
For CVRR-ES and VideoEval-Pro, a text-only LLM judge receives the question, reference answer, and generated answer, and produces one judgment at temperature 0. The CVRR-ES prompt requests a correct/incorrect label, a 0--5 rating, and a reason, with the reported metric based on the correctness label. Its rubric accepts less specific answers unless the question requires the omitted detail. VideoEval-Pro classifies answers as correct, incorrect, or not attempted. An answer is accepted when it contains the required information without contradicting the reference, while incorrect and not-attempted answers receive zero credit. Following the benchmarks' official evaluation protocols, we use GPT-3.5 for CVRR-ES and GPT-4o-0806 for VideoEval-Pro.

\FloatBarrier
\begin{table*}[!t]
\centering
\caption{\textbf{Performance comparison on \emph{InternVL3.5} without retuning.} We use the main experiments' hyperparameters. Accuracies are percentages, with relative accuracy normalized to the uncompressed baseline.}
\label{tab:comparison_internvl}
\papertablestyle
\setlength{\tabcolsep}{3pt}
\begin{adjustbox}{max width=\linewidth}
\begin{tabular}{@{}lc*{6}{c}@{}}
\toprule
Method & Ratio & MVBench & LongVB & MLVU & VideoMME & Mean accuracy & Relative accuracy \\
\midrule
\textbf{InternVL3.5-8B}~\cite{internvl3.5} & 100\% & 73.3 & 59.6 & 66.9 & 64.9 & 66.2 & 100.0\% \\
\midrule
\rowcolor{gray!6}
Random & 10\% & 60.0 & 55.9 & 60.3 & 57.6 & 58.4 & 88.2\% \\
VidCom2~\cite{vidcom2} & 10\% & 58.6 & 54.8 & 59.3 & 57.0 & 57.4 & 86.7\% \\
PruneVID~\cite{prunevid} & 10\% & 59.5 & 56.8 & 59.4 & 56.5 & 58.1 & 87.7\% \\
\textbf{KiToke (Ours)} & 10\% & \textbf{61.3} & \textbf{57.1} & \textbf{61.8} & \textbf{58.2} & \textbf{59.6} & \textbf{90.0\%} \\
\midrule
\rowcolor{gray!6}
Random & 1\% & 50.1 & 49.1 & 50.7 & 49.0 & 49.7 & 75.1\% \\
VidCom2~\cite{vidcom2} & 1\% & 50.0 & 49.5 & 48.0 & 48.0 & 48.9 & 73.8\% \\
PruneVID~\cite{prunevid} & 1\% & 51.4 & 49.8 & 49.6 & 47.7 & 49.6 & 75.0\% \\
\textbf{KiToke (Ours)} & 1\% & \textbf{53.2} & \textbf{51.9} & \textbf{53.3} & \textbf{51.2} & \textbf{52.4} & \textbf{79.2\%} \\
\bottomrule
\end{tabular}
\end{adjustbox}
\appendixanalysis{\textbf{InternVL3.5.} Table~\ref{tab:comparison_internvl} extends the evaluation to InternVL3.5-8B with InternViT~\cite{internvit}, a third visual-encoder family alongside SigLIP~\cite{siglip} in the LLaVA backbones and the SigLIP-2-based encoder~\cite{siglip2} in Qwen3-VL. Without hyperparameter retuning, KiToke outperforms VidCom2 and PruneVID at both 10\% and 1\% retention, preserving 90.0\% and 79.2\% of the uncompressed model's mean accuracy, respectively.}
\end{table*}

\begin{table*}[!t]
  \centering
  \caption{\textbf{Open-ended QA and cross-domain evaluation without retuning.} Acc. denotes accuracy (\%), averaged across 11 dimensions for CVRR-ES and across questions for the other benchmarks. Relative accuracy (\%) is normalized by the uncompressed LLaVA-OneVision-7B baseline.}
  \label{tab:open_ended_cross_domain}
  \papertablestyle
\begin{adjustbox}{max width=\linewidth}
  \begin{tabular}{@{}lc*{4}{w{c}{2.5em}w{c}{3.6em}}@{}}
  \toprule
  \multirow{2}{*}{Method} & \multirow{2}{*}{Ratio} &
  \multicolumn{2}{c}{EgoSchema} &
  \multicolumn{2}{c}{Video-MMMU} &
  \multicolumn{2}{c}{CVRR-ES} &
  \multicolumn{2}{c}{VideoEval-Pro} \\
\cmidrule(lr){3-4}\cmidrule(lr){5-6}\cmidrule(lr){7-8}\cmidrule(lr){9-10}
   & & Acc. & Relative & Acc. & Relative & Acc. & Relative & Acc. & Relative \\
  \midrule
  LLaVA-OV-7B & 100\% & 65.4 & 100\% & 32.5 & 100\% & 51.5 & 100\% & 28.0 & 100\% \\
  \midrule
  \rowcolor{gray!6}
  Random & 10\% & 62.8 & 96.0\% & 29.4 & 90.6\% & 40.6 & 78.9\% & 21.4 & 76.5\% \\
  FastVID~\cite{fastvid} & 10\% & 60.4 & 92.4\% & 22.0 & 67.7\% & 33.8 & 65.6\% & 8.9 & 31.8\% \\
  PruneVID~\cite{prunevid} & 10\% & 64.4 & 98.5\% & 28.1 & 86.5\% & 48.8 & 94.8\% & 25.3 & 90.4\% \\
  \textbf{KiToke (Ours)} & 10\% & \textbf{65.6} & \textbf{100.3\%} & \textbf{31.1} & \textbf{95.7\%} & \textbf{50.1} & \textbf{97.3\%} & \textbf{25.8} & \textbf{92.1\%} \\
  \midrule
  \rowcolor{gray!6}
  Random & 1\% & 55.4 & 84.7\% & 28.7 & 88.2\% & 35.4 & 68.8\% & 21.8 & 77.9\% \\
  FastVID~\cite{fastvid} & 1\% & 44.0 & 67.3\% & 16.1 & 49.5\% & 30.7 & 59.6\% & 6.2 & 22.1\% \\
  PruneVID~\cite{prunevid} & 1\% & 48.8 & 74.6\% & 25.2 & 77.5\% & 41.6 & 80.8\% & 21.6 & 77.1\% \\
  \textbf{KiToke (Ours)} & 1\% & \textbf{58.2} & \textbf{89.0\%} & \textbf{30.2} & \textbf{92.9\%} & \textbf{47.7} & \textbf{92.6\%} & \textbf{22.9} & \textbf{81.8\%} \\
  \bottomrule
  \end{tabular}
\end{adjustbox}
  \appendixanalysis{Table~\ref{tab:open_ended_cross_domain} shows that KiToke outperforms FastVID and PruneVID at both evaluated retention ratios across the open-ended and cross-domain benchmarks. These results extend the main comparisons beyond multiple-choice QA and support the transfer of the same compression strategy to different answer formats and domains.}
\end{table*}

\section{Additional Experimental Results\label{appendix:exp_results}}
\label{appendix:additional_experiments}

\subsection{Comparison with Diversity/Coverage-based Pruning\label{appendix:comparison_diversity}}

\begin{table*}[tp]
\centering
\caption{\textbf{Comparison with diversity/coverage-based pruning on LLaVA-Video.} We report four-benchmark mean accuracy (\%) under matched token budgets and normalize it by the uncompressed model's mean to obtain relative accuracy.}
\label{tab:comparison_diversity}
\papertablestyle
\begin{adjustbox}{max width=\linewidth}
\begin{tabular}{@{}lc*{6}{c}@{}}
\toprule
Method & \# Tokens & MVBench & LongVB & MLVU & VideoMME & Mean accuracy & Relative accuracy \\
\midrule
LLaVA-Video-7B & 10816 & 60.4 & 58.7 & 67.4 & 64.2 & 62.7 & 100.0\% \\
\midrule
k-center~\cite{k-center} & 4096 & 57.6 & 57.5 & 66.0 & 61.2 & 60.6 & 96.7\% \\
DivPrune~\cite{divprune} & 4096 & 57.9 & 57.8 & 66.7 & 61.1 & 60.9 & 97.1\% \\
CDPruner~\cite{cdpruner} & 4096 & 57.4 & \textbf{58.7} & 66.3 & 62.6 & 61.3 & 97.7\% \\
\textbf{KiToke (Ours)} & 4096 & \textbf{59.5} & 58.4 & \textbf{67.5} & \textbf{63.3} & \textbf{62.2} & \textbf{99.2\%} \\
\midrule
k-center~\cite{k-center} & 2048 & 55.3 & 55.7 & 62.1 & 58.6 & 57.9 & 92.3\% \\
DivPrune~\cite{divprune} & 2048 & 55.9 & 56.0 & 62.7 & 59.1 & 58.4 & 93.1\% \\
CDPruner~\cite{cdpruner} & 2048 & 55.7 & 56.5 & 63.0 & 60.5 & 58.9 & 94.0\% \\
\textbf{KiToke (Ours)} & 2048 & \textbf{58.3} & \textbf{58.2} & \textbf{66.5} & \textbf{61.2} & \textbf{61.1} & \textbf{97.4\%} \\
\midrule
k-center~\cite{k-center} & 1024 & 52.7 & 52.1 & 57.8 & 56.2 & 54.7 & 87.2\% \\
DivPrune~\cite{divprune} & 1024 & 53.4 & 52.9 & 58.6 & 56.8 & 55.4 & 88.4\% \\
CDPruner~\cite{cdpruner} & 1024 & 53.8 & 52.7 & 58.9 & 57.3 & 55.7 & 88.8\% \\
\textbf{KiToke (Ours)} & 1024 & \textbf{57.2} & \textbf{56.9} & \textbf{64.5} & \textbf{59.0} & \textbf{59.4} & \textbf{94.7\%} \\
\bottomrule
\end{tabular}
\end{adjustbox}
\end{table*}

To compare KiToke with other ways of selecting a representative token subset, Table~\ref{tab:comparison_diversity} evaluates diversity- and coverage-based pruning on LLaVA-Video. The alternatives include DivPrune~\cite{divprune} for pairwise-distance-based diversity and CDPruner~\cite{cdpruner} for conditional diversity through a DPP-based objective. We also evaluate classical $k$-center~\cite{k-center}, implemented through greedy farthest-point selection in the same feature space. All methods use the same post-pooling pipeline and matched output token budgets. Against these pruning methods, the complete KiToke pipeline, including distribution-aware selection and interval-aware merging, achieves the highest mean accuracy at each budget, with larger gains as the budget decreases.

\subsection{Selection with Fixed Merging}\label{appendix:selection_fixed_merging}

\begin{table*}[tp]
\centering
\caption{\textbf{Query-agnostic selection with fixed interval-aware merging on LLaVA-OneVision-7B at 1\% retention.} Columns report accuracy (\%) on four benchmarks and their unweighted mean. All selectors use KiToke's fixed repair and merging pipeline, while the KDE variants differ in the scope of density estimation for selection.}
\label{tab:selection_fixed_merging}
\papertablestyle
\begin{adjustbox}{max width=\linewidth}
\begin{tabular}{@{}l*{5}{c}@{}}
\toprule
Anchor selector & MVBench & LongVB & MLVU & VideoMME & Mean accuracy (\%) \\
\midrule
\rowcolor{gray!6}
Uniform sampling (Random) & 50.2 & 48.1 & 54.5 & 48.9 & 50.4 \\
k-center~\cite{k-center} & 50.0 & 47.7 & 54.5 & 48.2 & 50.1 \\
DivPrune~\cite{divprune} & 50.7 & 48.4 & 55.2 & 48.9 & 50.8 \\
FastVID~\cite{fastvid} & 50.4 & 48.0 & 54.8 & 48.4 & 50.4 \\
FLoC~\cite{floc} & 50.3 & 47.6 & 54.9 & 48.2 & 50.3 \\
Density-proportional sampling & 48.6 & 46.3 & 52.9 & 47.4 & 48.8 \\
Per-frame KDE, Inverse-density sampling & 50.9 & 48.4 & 55.1 & 48.8 & 50.8 \\
Interval KDE, Inverse-density sampling & 51.4 & 49.2 & 55.9 & 49.5 & 51.5 \\
\textbf{Full KDE, Inverse-density sampling (KiToke)} & \textbf{52.5} & \textbf{51.1} & \textbf{56.9} & \textbf{50.2} & \textbf{52.7} \\
\bottomrule
\end{tabular}
\end{adjustbox}
\end{table*}

The comparison in Table~\ref{tab:selection_fixed_merging} isolates the selector by giving each method the same downstream repair and merging pipeline. It provides the benchmark-level results for Table~\ref{tab:allocation_principles} on LLaVA-OneVision-7B at 1\% retention. All selectors receive the same visual tokens and budget $B=\lfloor\gamma N\rfloor$, and use KiToke's interval boundaries, empty-interval repair with $S_i$, L2 merge assignment, inverse-density merge weights, and output ordering. Uniform sampling, density-proportional sampling ($q_i\propto D_i$), and all inverse-density variants ($q_i\propto S_i$) also share the pivotal sampler and inclusion-probability conversion in Equation~\ref{eq:inclusion_probabilities}.

To examine the scope of density estimation, we compare inverse-density selection based on per-frame KDE, interval KDE, and full KDE. These estimate density within each frame, within each detected temporal interval, and across all video tokens, respectively. Density-proportional sampling also uses full-video densities, while repair and merging use the same full-video scores $S_i=1/D_i$ across all variants.

Full KDE with inverse-density sampling achieves the highest accuracy on all four benchmarks, reaching a mean of 52.7\%. This exceeds uniform sampling by 2.3 percentage points, DivPrune by 1.9 points, and FLoC by 2.4 points, whereas density-proportional sampling reaches 48.8\%, below uniform sampling's 50.4\%. With repair and merging held fixed, the comparison supports giving sparsely represented tokens greater selection weight at this retention ratio.

Full KDE also outperforms per-frame and interval KDE on every benchmark, improving mean accuracy by 1.9 and 1.2 percentage points, respectively. These results support estimating redundancy over the full video to account for visual patterns recurring across distant frames and intervals.

\subsection{Temporal Boundaries and Merge Weights}\label{appendix:temporal_merge_ablation}

\begin{table*}[tp]
\centering
\caption{\textbf{Ablation of temporal boundaries and merge weights on LLaVA-OneVision-7B at 1\% retention.} Both ``Matched $L$'' controls and uniform intervals use the same number of intervals as KiToke for each video after budget coarsening. The $c_t$-ranked control places boundaries at the $L-1$ transitions with the largest adjacent-frame change scores $c_t$, without using the local-increase signal. Accuracy columns report percentages, with an unweighted mean across the four benchmarks. Repair is the mean percentage of anchors replaced per question, including questions requiring no repair. Best accuracies are highlighted in bold.}
\label{tab:temporal_merge_ablation}
\papertablestyle
\begin{adjustbox}{max width=\linewidth}
\begin{tabular}{@{}ll*{6}{c}@{}}
\toprule
Boundary rule & Merge weights & MVBench & LongVB & MLVU & VideoMME & \shortstack{Mean accuracy (\%)} & \shortstack{Repair (\%)} \\
\midrule
Uniform intervals & Inverse density & 51.2 & 49.3 & 56.2 & 49.0 & 51.4 & 1.66 \\
FastVID-style, matched $L$ & Inverse density & 51.9 & 49.8 & 56.5 & 49.5 & 51.9 & 1.75 \\
$c_t$-ranked, matched $L$ & Inverse density & 52.1 & 50.0 & 56.3 & 49.1 & 51.9 & 1.73 \\
\midrule
KiToke & Uniform & 52.1 & 50.6 & 56.6 & 49.6 & 52.2 & 1.72 \\
\textbf{KiToke} & \textbf{Inverse density} & \textbf{52.5} & \textbf{51.1} & \textbf{56.9} & \textbf{50.2} & \textbf{52.7} & 1.72 \\
\bottomrule
\end{tabular}
\end{adjustbox}
\end{table*}

We next separate the effect of temporal boundary placement from that of merge weighting, using LLaVA-OneVision-7B at 1\% retention. In Table~\ref{tab:temporal_merge_ablation}, all variants start from the same anchors sampled with full-video inverse-density scores for each video and seed. The boundary comparisons keep the repair rule, L2 merge assignment, inverse-density merge weights, output order, and token budget fixed.

To compare boundary placement at the same granularity, each variant uses KiToke's interval count $L$ after budget coarsening. Uniform intervals divide the frame sequence into $L$ contiguous parts whose lengths differ by at most one frame. The FastVID-style control selects the $L-1$ transitions with the lowest FastVID frame similarity, whereas the $c_t$-ranked control selects the $L-1$ largest KiToke change scores, breaking ties chronologically. Because the boundaries determine which intervals need repair, we also report the fraction of anchors replaced.

The merge-weight comparison reuses KiToke's boundaries, repaired anchors, and token-to-anchor assignments. Only the averaging weights change: uniform weighting assigns weight one to the anchor and each assigned token, while inverse-density weighting uses their full-video scores $S_i$.

\noindent\textbf{Boundary placement.}
With inverse-density merge weights fixed, KiToke reaches 52.7\% mean accuracy, compared with 51.4\% for uniform intervals and 51.9\% for both FastVID-style and $c_t$-ranked boundaries, improving all four benchmarks. Matching the interval count shows that the gain depends on where boundaries are placed, beyond the benefit of restricting merging to temporal neighborhoods. The advantage over $c_t$ ranking further supports using local increases to identify boundaries. Across these variants, repair replaces 1.66--1.75\% of the sampled anchors.

\noindent\textbf{Merge weights.}
With KiToke's boundaries, repaired anchors, and token-to-anchor assignments fixed, inverse-density merge weights increase mean accuracy from 52.2\% to 52.7\%. The gain appears on all four benchmarks, isolating the contribution of weighting the merged features.

\subsection{Efficiency on Qwen3-VL\label{appendix:efficiency_qwen3_vl}}

\begin{table*}[tp]
\centering
\caption{\textbf{Prefill latency on Qwen3-VL-8B.} Compression time includes exact density estimation, token selection, and interval-aware merging. Total latency is the sum of compression time and LLM prefill time, averaged over the same 100 VideoMME samples at batch size 1 on a single NVIDIA A100-80GB GPU. The memory column reports total peak GPU memory over the same measurement run.}
\label{tab:efficiency_qwen3_vl}
\papertablestyle
\begin{adjustbox}{max width=\linewidth}
\begin{tabular}{@{}lccccc@{}}
\toprule
Method & \# Video Tokens & Comp. Time (s) & LLM prefill (s) & Total Latency (s) & Total peak GPU memory (GiB) \\
\midrule
Vanilla & 32K & 0 & 19.62 & 19.62 & 44.89 \\
KiToke (Ours) & 0.32K (1\%) & 0.55 & 0.32 & \textbf{0.87 (22.6$\times$)} & \textbf{38.15} \\
\midrule
Vanilla & 16K & 0 & 4.32 & 4.32 & 25.80  \\
KiToke (Ours) & 0.16K (1\%) & 0.18 & 0.22 & \textbf{0.40 (10.8$\times$)} & \textbf{22.88}  \\
\bottomrule
\end{tabular}
\end{adjustbox}
\end{table*}

The Qwen3-VL-8B measurements in Table~\ref{tab:efficiency_qwen3_vl} examine whether compression remains beneficial with larger candidate sequences, using exact density estimation and the timing protocol in Appendix~\ref{appendix:latency_protocol}. At 32K video tokens, KiToke reduces LLM prefill time from 19.62 to 0.32\,s, with compression taking 0.55\,s. The combined latency of 0.87\,s gives a 22.6$\times$ speedup over uncompressed prefill. At 16K tokens, combined latency falls from 4.32 to 0.40\,s, a 10.8$\times$ speedup.

The following experiment examines whether approximate nearest-neighbor density estimation~\cite{ann-kde} can further reduce the compression cost at 32K candidate tokens.

\subsection{Approximate Density Estimation}\label{appendix:ann_kde}

\begin{table*}[tp]
\centering
\caption{\textbf{Exact versus approximate density estimation on Qwen3-VL-8B with 32K candidate tokens.} Density-estimation costs precede ratio-specific sampling and apply at both retention ratios. ``Other compression'' excludes density estimation, whereas ``Total'' sums density estimation, other compression operations, and LLM prefill. Mean accuracy averages the four benchmarks (\%).}
\label{tab:ann_kde}
\papertablestyle
\begin{adjustbox}{max width=\linewidth}
\begin{tabular}{@{}lc*{7}{c}@{}}
\toprule
 & & \multicolumn{3}{c}{Density estimation} & \multicolumn{3}{c}{Latency (s)} & \\
\cmidrule(lr){3-5}\cmidrule(lr){6-8}
Estimator & Retention & TFLOPs & \shortstack{Incremental peak\\GPU allocation (GiB)} & Time (s) & \shortstack{Other\\compression} & LLM prefill & Total & Mean accuracy \\
\midrule
Uncompressed & 100\% & -- & -- & -- & -- & 19.62 & 19.62 & 68.5 \\
\midrule
Exact & 10\% & 7.75 & 15.88 & 0.51 & 0.06 & 1.03 & 1.60 & 64.0 \\
ANN & 10\% & 0.92 & 0.64 & 0.06 & 0.06 & 1.03 & 1.15 & 63.7 \\
\addlinespace[2pt]
Exact & 1\% & 7.75 & 15.88 & 0.51 & 0.04 & 0.32 & 0.87 & 54.1 \\
ANN & 1\% & 0.92 & 0.64 & 0.06 & 0.04 & 0.32 & 0.42 & 53.9 \\
\bottomrule
\end{tabular}
\end{adjustbox}
\end{table*}
Exact density estimation evaluates similarities across the full token population, making its cost grow quadratically with the number of candidates. We therefore compare it with an ANN-based alternative on Qwen3-VL-8B at the 32K candidate-token setting. This variant retrieves neighbors through ANN search~\cite{ann-kde}, sums their Gaussian contributions, and uses the resulting scores in the same selection and merging stages. For $k$ retrieved neighbors and feature dimension $d$, kernel aggregation costs $O(Nkd)$, with additional costs for index construction and retrieval. Table~\ref{tab:ann_kde} measures the empirical accuracy--cost trade-off of this truncated estimator against exact density estimation.

Alongside accuracy, we report density-estimation compute, memory, and runtime, together with prefill latency. Exact-estimation runtime covers pairwise distance computation, kernel evaluation, and reduction. For ANN, runtime also includes index construction and retrieval. Both use the shared timing protocol on 100 VideoMME samples at batch size 1 on an A100-80GB. This experiment uses 32K candidate tokens, compared with up to 16K in the main Qwen3-VL evaluation.

ANN reduces density-estimation runtime by 8.5$\times$, FLOPs by 8.4$\times$, and incremental peak GPU allocation by 24.8$\times$, with little loss of accuracy. At 1\% retention, total latency decreases from 0.87 to 0.42\,s while accuracy falls by 0.2 percentage points. At 10\%, latency decreases from 1.60 to 1.15\,s with accuracy down 0.3 percentage points.

The ANN implementation uses a PyTorch CUDA IVF-Flat index with squared Euclidean distance, $k=128$, 256 coarse lists, and eight probed lists, including the query's own list. For each video, we train the index for five k-means iterations on up to 8,192 evenly spaced tokens. Index construction, retrieval, and kernel aggregation all run on the GPU, with the self contribution retained and the truncated density sum left unscaled. This configuration was selected through an unlabeled pilot on four EgoSchema videos at each retention ratio, comparing $k\in\{32,64,128,256\}$. The selected configuration gives a density Spearman correlation of 0.9722.

Density-estimation FLOPs count additions and multiplications as one operation each and fused multiply--adds as two, excluding transcendental functions, comparisons, and integer operations. Incremental peak GPU allocation is measured relative to allocated memory at entry to density estimation, after the caller's feature conversion, and reported in GiB ($1\,\mathrm{GiB}=2^{30}$ bytes). Timing uses five warmups, then sequential runs on one GPU.

\subsection{Long-Video Inference with a Fixed Output Budget}
\label{sec:fixed_budget}

\begin{table*}[tp]
\centering
\caption{\textbf{Increasing input-frame coverage under a fixed LLM visual-token budget.}}
\label{tab:length_extrapolation}
\papertablestyle
\begin{adjustbox}{max width=\linewidth}
\begin{tabular}{@{}lccc*{4}{c}@{}}
\toprule
\multirow{2}{*}{Method} & \multirow{2}{*}{\# Frames} & \multirow{2}{*}{Retention ratio $\gamma$} & \multirow{2}{*}{\# Tokens} & \multicolumn{4}{c}{VideoMME} \\
\cmidrule(lr){5-8}
 &  &  &  & Overall & Short & Medium & Long \\
\midrule
Vanilla & 32 & 100\% & 6272 & 58.4 & 70.3 & 56.3 & 48.6 \\
\addlinespace[2pt]
KiToke (Ours) & 64 & 50\% & 6272 & 59.6 & 71.9 & 57.7 & 49.3 \\
KiToke (Ours) & 128 & 25\% & 6272 & 60.5 & 71.9 & 58.4 & 51.1 \\
KiToke (Ours) & 320 & 10\% & 6272 & 61.6 & 72.5 & 59.9 & 52.3 \\
\bottomrule
\end{tabular}
\end{adjustbox}
\end{table*}

\begin{table*}[tp]
\centering
\caption{\textbf{Ablation study of selection methods using inverse-density scores.} Values for stochastic methods are averaged over 10 random seeds, with 95\% confidence intervals across seeds shown in parentheses.}
\label{tab:token_selection_ablation}
\papertablestyle
\begin{adjustbox}{max width=\linewidth}
\begin{tabular}{@{}lccccc@{}}
\toprule
Method & MVBench & LongVB & MLVU & VideoMME & Mean accuracy (\%) \\
\midrule
Top-K & 51.1 & 48.9 & 55.5 & 49.1 & 51.2  \\
Multinomial Sampling & 52.5 (±0.26) & 49.7 (±0.51) & 56.6 (±0.38) & 50.2 (±0.31) & 52.3 (±0.23) \\
\textbf{Pivotal Sampling} & \textbf{52.5 (±0.25)} & \textbf{51.1 (±0.25)} & \textbf{56.9 (±0.29)} & \textbf{50.2 (±0.29)} & \textbf{52.7 (±0.16)} \\
\bottomrule
\end{tabular}
\end{adjustbox}
\end{table*}

\begin{table*}[tp]
\centering
\caption{\textbf{Empty-interval repair at 1\% retention on LLaVA-OneVision.} Accuracy and repair statistics are percentages, while accuracy gain is measured in percentage points (pp). Repair frequency is the fraction of questions requiring repair. Replaced anchors is the mean fraction replaced per question, including those needing no repair.}
\label{tab:repair_ablation}
\papertablestyle
\begin{adjustbox}{max width=\linewidth}
\begin{tabular}{@{}l*{5}{c}@{}}
\toprule
Benchmark & With repair & Without repair & Accuracy gain (pp) & Repair frequency & Replaced anchors \\
\midrule
MVBench & 52.525 & 52.500 & +0.025 & 2.55 & 0.05 \\
LongVideoBench & 51.085 & 51.010 & +0.075 & 48.69 & 2.14 \\
MLVU & 56.865 & 56.581 & +0.284 & 62.14 & 2.62 \\
VideoMME & 50.185 & 50.074 & +0.111 & 71.00 & 3.27 \\
\bottomrule
\end{tabular}
\end{adjustbox}
\end{table*}

A fixed LLM input budget allows longer videos to be processed without increasing the number of tokens entering the language model. As the number of candidate tokens $N$ grows, however, the retained fraction $\gamma=B/N$ must fall to keep the output budget $B$ fixed.

Table~\ref{tab:length_extrapolation} evaluates this setting by increasing the LLaVA-OneVision input from 32 to 320 frames while keeping the LLM input at 6272 visual tokens. As retention falls from 100\% to 10\%, VideoMME accuracy rises from 58.4 to 61.6 overall and from 48.6 to 52.3 on long videos. The additional temporal coverage therefore improves accuracy even though the LLM receives the same number of visual tokens.

\subsection{Inverse-density Sampling}

To examine how the inverse-density scores should be used, Table~\ref{tab:token_selection_ablation} compares deterministic top-$K$, multinomial sampling, and pivotal sampling with the scores held fixed. Top-$K$ gives the lowest accuracy, consistent with its tendency to omit dense regions by retaining only the highest-scoring tokens. Stochastic sampling keeps tokens from those regions eligible for selection, with pivotal sampling~\cite{deville1998splitting} achieving slightly higher accuracy and lower variance than standard multinomial sampling.

\subsection{Effect of Empty-interval Repair}\label{appendix:repair_ablation}

We evaluate empty-interval repair on LLaVA-OneVision at 1\% retention, starting both variants from the same sampled anchors. Intervals containing no anchors contribute no features unless repaired, and this repair raises the four-benchmark mean accuracy by 0.124 percentage points in Table~\ref{tab:repair_ablation}. Although 39.38\% of the 10,211 questions require repair, only 1.72\% of anchors are replaced overall. The procedure therefore ensures interval coverage through changes to a small fraction of the sampled anchors.

\subsection{Seed Stability\label{appendix:seed_stability}}

\begin{table*}[tp]
\centering
\caption{\textbf{Stochastic-method statistics across backbones and retention ratios.} We report per-benchmark and mean accuracy (\%) for Random and KiToke on LLaVA-OneVision, LLaVA-Video, and Qwen3-VL. Parentheses show 95\% confidence intervals across 10 seeds.}
\label{tab:stochastic_ci_template}
\papertablestyle
\begin{adjustbox}{max width=\linewidth}
\begin{tabular}{@{}clccccc@{}}
\toprule
Retention & Method & MVBench & LongVB & MLVU & VideoMME & Mean accuracy (\%) \\
\midrule
\multicolumn{7}{@{}l}{\textbf{LLaVA-OneVision}} \\
\addlinespace[2pt]
\multirow{2}{*}{35\%} & Random & 56.4 (±0.20) & 57.0 (±0.24) & 62.3 (±0.20) & 57.8 (±0.24) & 58.4 (±0.15) \\
 & KiToke (Ours) & 58.2 (±0.12) & 57.5 (±0.15) & 62.9 (±0.14) & 58.0 (±0.13) & 59.1 (±0.09) \\
\addlinespace[2pt]
\multirow{2}{*}{25\%} & Random & 56.7 (±0.25) & 57.1 (±0.30) & 61.9 (±0.25) & 56.7 (±0.29) & 58.1 (±0.19) \\
 & KiToke (Ours) & 58.4 (±0.14) & 57.4 (±0.17) & 63.4 (±0.17) & 58.3 (±0.19) & 59.4 (±0.10) \\
\addlinespace[2pt]
\multirow{2}{*}{15\%} & Random & 56.0 (±0.30) & 56.1 (±0.36) & 61.1 (±0.29) & 56.1 (±0.35) & 57.3 (±0.23) \\
 & KiToke (Ours) & 57.6 (±0.16) & 56.5 (±0.19) & 62.5 (±0.20) & 57.7 (±0.22) & 58.6 (±0.12) \\
\addlinespace[2pt]
\multirow{2}{*}{10\%} & Random & 54.2 (±0.36) & 54.7 (±0.44) & 59.7 (±0.34) & 54.7 (±0.41) & 55.8 (±0.27) \\
 & KiToke (Ours) & 57.2 (±0.19) & 57.4 (±0.21) & 62.1 (±0.23) & 56.0 (±0.25) & 58.2 (±0.13) \\
\addlinespace[2pt]
\multirow{2}{*}{5\%} & Random & 53.0 (±0.44) & 53.9 (±0.54) & 58.2 (±0.40) & 52.5 (±0.49) & 54.4 (±0.32) \\
 & KiToke (Ours) & 55.2 (±0.22) & 54.9 (±0.23) & 60.8 (±0.26) & 54.1 (±0.26) & 56.2 (±0.15) \\
\addlinespace[2pt]
\multirow{2}{*}{1\%} & Random & 48.0 (±0.52) & 46.9 (±0.64) & 54.6 (±0.47) & 49.0 (±0.58) & 49.6 (±0.38) \\
 & KiToke (Ours) & 52.5 (±0.25) & 51.1 (±0.25) & 56.9 (±0.29) & 50.2 (±0.29) & 52.7 (±0.16) \\
\midrule
\multicolumn{7}{@{}l}{\textbf{LLaVA-Video}} \\
\addlinespace[2pt]
\multirow{2}{*}{35\%} & Random & 57.7 (±0.22) & 56.2 (±0.28) & 63.8 (±0.22) & 61.1 (±0.26) & 59.7 (±0.16) \\
 & KiToke (Ours) & 59.5 (±0.15) & 58.3 (±0.19) & 67.3 (±0.16) & 63.0 (±0.19) & 62.0 (±0.11) \\
\addlinespace[2pt]
\multirow{2}{*}{25\%} & Random & 56.9 (±0.27) & 55.9 (±0.34) & 61.5 (±0.27) & 60.1 (±0.31) & 58.6 (±0.20) \\
 & KiToke (Ours) & 58.7 (±0.19) & 58.2 (±0.24) & 66.5 (±0.19) & 62.1 (±0.23) & 61.4 (±0.14) \\
\addlinespace[2pt]
\multirow{2}{*}{15\%} & Random & 55.3 (±0.33) & 55.9 (±0.42) & 59.6 (±0.32) & 58.7 (±0.38) & 57.4 (±0.24) \\
 & KiToke (Ours) & 58.2 (±0.23) & 58.0 (±0.29) & 66.1 (±0.23) & 61.0 (±0.27) & 60.8 (±0.17) \\
\addlinespace[2pt]
\multirow{2}{*}{10\%} & Random & 54.1 (±0.40) & 55.1 (±0.51) & 59.2 (±0.38) & 57.8 (±0.46) & 56.6 (±0.29) \\
 & KiToke (Ours) & 57.2 (±0.28) & 57.0 (±0.35) & 65.1 (±0.27) & 59.1 (±0.33) & 59.6 (±0.21) \\
\addlinespace[2pt]
\multirow{2}{*}{5\%} & Random & 52.9 (±0.48) & 52.7 (±0.61) & 57.3 (±0.46) & 55.4 (±0.55) & 54.6 (±0.35) \\
 & KiToke (Ours) & 55.6 (±0.34) & 55.7 (±0.43) & 62.5 (±0.32) & 57.4 (±0.40) & 57.8 (±0.25) \\
\addlinespace[2pt]
\multirow{2}{*}{1\%} & Random & 47.6 (±0.58) & 48.9 (±0.74) & 53.1 (±0.55) & 49.5 (±0.66) & 49.8 (±0.42) \\
 & KiToke (Ours) & 49.0 (±0.41) & 51.9 (±0.52) & 55.8 (±0.38) & 52.3 (±0.48) & 52.3 (±0.30) \\
\midrule
\multicolumn{7}{@{}l}{\textbf{Qwen3-VL}} \\
\addlinespace[2pt]
\multirow{2}{*}{35\%} & Random & 66.6 (±0.25) & 61.0 (±0.32) & 69.3 (±0.24) & 66.3 (±0.29) & 65.8 (±0.18) \\
 & KiToke (Ours) & 67.1 (±0.17) & 62.4 (±0.22) & 69.4 (±0.17) & 67.0 (±0.20) & 66.5 (±0.12) \\
\addlinespace[2pt]
\multirow{2}{*}{25\%} & Random & 64.5 (±0.30) & 61.0 (±0.39) & 68.6 (±0.29) & 64.9 (±0.35) & 64.8 (±0.22) \\
 & KiToke (Ours) & 65.2 (±0.21) & 61.5 (±0.27) & 69.5 (±0.21) & 65.8 (±0.25) & 65.5 (±0.15) \\
\addlinespace[2pt]
\multirow{2}{*}{15\%} & Random & 61.0 (±0.36) & 58.7 (±0.47) & 66.6 (±0.33) & 62.0 (±0.41) & 62.1 (±0.27) \\
 & KiToke (Ours) & 62.9 (±0.25) & 60.4 (±0.32) & 67.6 (±0.25) & 64.1 (±0.29) & 63.8 (±0.18) \\
\addlinespace[2pt]
\multirow{2}{*}{10\%} & Random & 57.6 (±0.43) & 57.1 (±0.56) & 65.3 (±0.40) & 61.7 (±0.49) & 60.5 (±0.32) \\
 & KiToke (Ours) & 60.5 (±0.30) & 58.6 (±0.38) & 67.0 (±0.30) & 62.8 (±0.34) & 62.2 (±0.21) \\
\addlinespace[2pt]
\multirow{2}{*}{5\%} & Random & 53.1 (±0.52) & 54.7 (±0.67) & 60.3 (±0.49) & 59.4 (±0.59) & 56.9 (±0.38) \\
 & KiToke (Ours) & 57.1 (±0.36) & 56.7 (±0.46) & 63.7 (±0.36) & 59.9 (±0.42) & 59.3 (±0.27) \\
\addlinespace[2pt]
\multirow{2}{*}{1\%} & Random & 45.9 (±0.62) & 48.2 (±0.80) & 53.8 (±0.58) & 52.7 (±0.70) & 50.2 (±0.45) \\
 & KiToke (Ours) & 48.2 (±0.43) & 50.7 (±0.55) & 56.3 (±0.40) & 53.6 (±0.49) & 52.2 (±0.32) \\
\bottomrule
\end{tabular}
\end{adjustbox}
\appendixanalysis{In Table~\ref{tab:stochastic_ci_template}, confidence intervals generally narrow as retention increases, reflecting lower sensitivity to the sampled tokens when more are retained. At a given retention ratio and backbone, KiToke tends to achieve higher mean accuracy than Random with comparable or narrower intervals. Although intervals are often somewhat wider on long-form benchmarks, the mean accuracy gap between Random and KiToke at $5\%$ and $1\%$ retention remains large relative to these seed-level intervals.}
\end{table*}

Because KiToke selects anchors stochastically, we examine how predictions vary across random seeds. Table~\ref{tab:token_selection_ablation} compares sampling strategies at $1\%$ retention on LLaVA-OneVision with $95\%$ confidence intervals across seeds. Table~\ref{tab:stochastic_ci_template} extends this analysis to Random and KiToke on LLaVA-OneVision, LLaVA-Video, and Qwen3-VL from $35\%$ to $1\%$ retention, reporting intervals for each benchmark and the four-benchmark mean.

We also examine prediction consistency for individual questions on LLaVA-OneVision at $10\%$ retention over the same four benchmarks and 10 seeds. Predictions are identical across all 10 seeds for $74.6\%$ of samples, and at most one prediction differs for $82.2\%$. On average, $9.4/10$ runs agree with the majority prediction.

\FloatBarrier
\subsection{Kernel and Merge-Metric Sensitivity}\label{appendix:kernel_sensitivity}

\begin{table*}[tp]
\centering
\caption{\textbf{Kernel sensitivity at 1\% retention.} LLaVA-OneVision-7B. Only the density kernel changes.}
\label{tab:kernel_sensitivity_full}
\papertablestyle
\begin{adjustbox}{max width=\linewidth}
\begin{tabular}{@{}l*{5}{c}@{}}
\toprule
Density kernel & MVBench & LongVB & MLVU & VideoMME & Mean accuracy (\%) \\
\midrule
Gaussian & 52.5 & 51.1 & 56.9 & 50.2 & 52.7 \\
Laplacian & 52.8 & 50.8 & 55.3 & 50.4 & 52.3 \\
Angular Gaussian & 52.2 & 50.7 & 56.3 & 50.1 & 52.3 \\
Rescaled cosine & 50.3 & 48.5 & 54.2 & 48.9 & 50.5 \\
\bottomrule
\end{tabular}
\end{adjustbox}
\end{table*}
We examine whether the results depend on the particular density kernel or merge-assignment metric, using LLaVA-OneVision-7B at 1\% retention on the four main benchmarks. For kernel sensitivity, all stages except the density kernel remain fixed. Letting $r_{ij}=\|\mathbf{x}_i-\mathbf{x}_j\|_2$ and $c_{ij}=\cos(\mathbf{x}_i,\mathbf{x}_j)$, we compare Gaussian $\exp(-r_{ij}^2/\alpha)$, Laplacian $\exp(-r_{ij}/\alpha_L)$, angular Gaussian $\exp(-(1-c_{ij})/\alpha_A)$, and rescaled cosine $(1+c_{ij})/2$. Gaussian uses the default $\alpha=800$, and the other bandwidths are calibrated to match its median off-diagonal kernel value.

The calibration protocol uses visual features from the first 16 EgoSchema test videos, disjoint from the four evaluation benchmarks. It samples 100,000 off-diagonal token pairs per video with replacement (1.6 million in total), with no QA labels. The calibrated bandwidths are $\alpha_L\approx10.481$ and $\alpha_A\approx0.074071$, fixed across benchmarks.

Table~\ref{tab:kernel_sensitivity_full} shows similar mean accuracy for Gaussian at 52.7\% and for Laplacian and angular Gaussian at 52.3\%. This 0.4-point gap supports robustness across the three kernels, with angular Gaussian retaining similar accuracy using feature directions alone. Rescaled cosine performs less well, falling 2.2 percentage points below Gaussian. Because an orthogonal token contributes $0.5$ under rescaled cosine, the accumulated contributions of many such tokens can dilute density contrasts.

To assess sensitivity to the merge metric, we vary how unselected tokens are assigned to anchors while holding the Gaussian density estimate, selected anchors, temporal intervals, and averaging weights fixed. Maximizing Gaussian similarity is equivalent to minimizing squared Euclidean distance, so the assignment can be computed without evaluating the exponential. Table~\ref{tab:merge_metric_full} shows an accuracy difference of 0.2 percentage points between cosine and Euclidean assignment.

\begin{table*}[tp]
\centering
\caption{\textbf{Merge-assignment sensitivity at 1\% retention.} Results use LLaVA-OneVision-7B with Gaussian density fixed.}
\label{tab:merge_metric_full}
\papertablestyle
\begin{adjustbox}{max width=\linewidth}
\begin{tabular}{@{}l*{5}{c}@{}}
\toprule
Assignment metric & MVBench & LongVB & MLVU & VideoMME & Mean accuracy (\%) \\
\midrule
Cosine & 52.5 & 51.2 & 56.3 & 49.9 & 52.5 \\
Gaussian / nearest $L_2$ & 52.5 & 51.1 & 56.9 & 50.2 & 52.7 \\
\bottomrule
\end{tabular}
\end{adjustbox}
\end{table*}

\subsection{Hyperparameter Sensitivity}\label{appendix:hyperparameter_sensitivity}

\begin{figure*}[tp]
  \centering
  \begin{subfigure}[t]{0.32\textwidth}
    \centering
    \includegraphics[width=\linewidth]{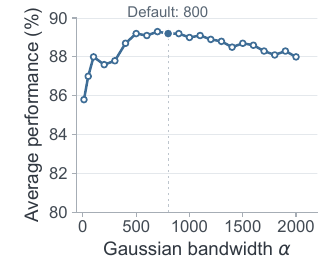}
    \caption{Gaussian bandwidth $\alpha$}
    \label{fig:bandwidth_panel}
  \end{subfigure}
  \hfill
  \begin{subfigure}[t]{0.32\textwidth}
    \centering
    \includegraphics[width=\linewidth]{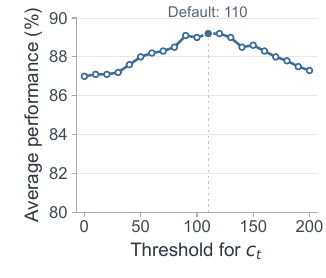}
    \caption{Change threshold $\tau_c$}
    \label{fig:change_threshold_panel}
  \end{subfigure}
  \hfill
  \begin{subfigure}[t]{0.32\textwidth}
    \centering
    \includegraphics[width=\linewidth]{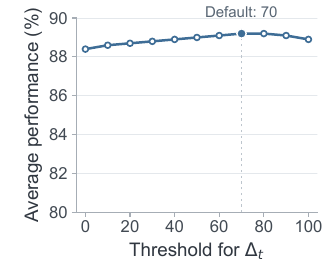}
    \caption{Absolute-increase threshold $\tau_\Delta$}
    \label{fig:absolute_threshold_panel}
  \end{subfigure}

\caption{\textbf{Hyperparameter ablations on LLaVA-OneVision-7B at 1\% visual-token retention.}
  Panels (a)--(c) vary Gaussian bandwidth $\alpha$, the threshold on adjacent-frame change $c_t$, and the threshold on its absolute increase $\Delta_t$, respectively. Filled markers identify the defaults $\alpha=800$ and $(\tau_c,\tau_\Delta)=(110,70)$.
  The vertical axis is mean accuracy over MVBench, LongVideoBench, MLVU, and VideoMME without subtitles, normalized by the uncompressed LLaVA-OneVision-7B mean (100\%).}
\label{fig:thres_ablation}
\end{figure*}

We evaluate sensitivity around the defaults calibrated in Appendix~\ref{appendix:hyperparameter_calibration}, using LLaVA-OneVision-7B at 1\% retention.

\noindent\textbf{Gaussian bandwidth.}
Panel~\subref{fig:bandwidth_panel} of Figure~\ref{fig:thres_ablation} shows that accuracy remains stable over a broad range of $\alpha$ around the calibrated default. Performance declines at the extremes: a very small $\alpha$ captures only near-neighbor redundancy, whereas a very large $\alpha$ makes token similarities nearly uniform.

\noindent\textbf{Temporal thresholds.}
Panels~\subref{fig:change_threshold_panel} and~\subref{fig:absolute_threshold_panel} of Figure~\ref{fig:thres_ablation} vary the thresholds on $c_t$ and $\Delta_t$. The calibrated defaults $(\tau_c,\tau_\Delta)=(110,70)$ perform well, with accuracy remaining robust to moderate changes in either threshold.

\subsection{Model Size}\label{appendix:model_size}

\begin{figure}[tp]
  \centering
  \includegraphics[width=\linewidth]{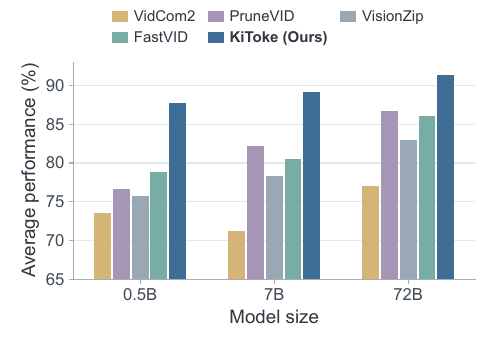}
  \caption{\textbf{Model-size ablation at 1\% visual-token retention.}
  Comparison at 0.5B, 7B, and 72B model sizes.
  The vertical axis reports normalized average accuracy: $100\bar A_{\mathrm{compressed}}/\bar A_{\mathrm{uncompressed}}$, where $\bar A$ is the mean accuracy over MVBench, LongVideoBench, MLVU, and VideoMME without subtitles. Each compressed result is normalized by the uncompressed result for the same model variant.}
  \label{fig:model_size_ablation}
\end{figure}

Figure~\ref{fig:model_size_ablation} extends the principal 7B/8B evaluations to 0.5B and 72B variants. KiToke consistently outperforms the prior training-free baselines at these sizes, with the largest gains on the 0.5B model, where compression has a greater effect on accuracy.

\subsection{Composition with Query-Aware Pruning}\label{appendix:query_composition}

\begin{table*}[tp]
\centering
\caption{\textbf{Query-agnostic and query-aware compression with a matched 5\% final budget.} LLaVA-OneVision-7B. Early-layer token counts differ, so final-budget matching does not equalize computation.}
\label{tab:query_composition_full}

\papertablestyle
\begin{adjustbox}{max width=\linewidth}
\begin{tabular}{@{}lcc*{5}{c}@{}}
\toprule
Method & Pre-LLM & Final & MVBench & LongVB & MLVU & VideoMME & Mean accuracy (\%) \\
\midrule
FastV & 100\% & 5\% & 49.9 & 48.3 & 55.4 & 51.5 & 51.3 \\
PruneVID (full) & 10\% & 5\% & 55.2 & 52.4 & 59.9 & 53.9 & 55.3 \\
FastVID + FastV & 10\% & 5\% & 54.0 & 52.2 & 56.7 & 53.7 & 54.2 \\
KiToke & 5\% & 5\% & 55.2 & 54.9 & 60.8 & 54.1 & 56.2 \\
KiToke + FastV & 10\% & 5\% & 56.0 & 55.0 & 60.6 & 55.6 & 56.8 \\
\bottomrule
\end{tabular}
\end{adjustbox}
\end{table*}
To examine both comparison and composition with query-aware methods, we evaluate KiToke, FastV~\cite{fastv}, full PruneVID, FastVID + FastV, and KiToke + FastV on LLaVA-OneVision-7B. All five pipelines finish at 5\% of the original visual-token count, but reach this budget at different stages. FastV prunes inside the LLM, whereas the hybrid methods first compress to 10\% before the LLM and then retain half of those tokens. Here PruneVID uses both its pre-LLM and query-aware intra-LLM components, extending the pre-LLM-only comparison in Table~\ref{tab:comparison_llava_ov}. All ratios are relative to the original token count.

KiToke alone achieves 56.2\% mean accuracy, exceeding FastV at 51.3\%, full PruneVID at 55.3\%, and FastVID + FastV at 54.2\%. Under the same 10\%-to-5\% schedule, KiToke + FastV reaches 56.8\%, outperforming FastVID + FastV by 2.6 accuracy percentage points. Its 0.6-point gain over KiToke alone comes with more tokens processed in the early LLM layers. These results show that query-aware pruning can complement KiToke's pre-LLM compression.

All composition experiments use 32 frames and batch size 1, with greedy decoding and at most 16 new tokens. The FastV-based rows share pruning-layer parameter $K=10$: FastV alone retains 5\% of its visual input, while each hybrid retains 50\% after pre-LLM compression to 10\%.

\subsection{Adding a Frame-global Attention Prior\label{appendix:global_attention_prior}}

\begin{table*}[tp]
\centering
\caption{\textbf{Adding a frame-global attention prior on \emph{LLaVA-OneVision}.} ``KiToke + global attn.'' adds pooling-token attention to the inverse-density score before sampling. Accuracies are percentages, and relative accuracy is normalized by the uncompressed mean.}
\label{tab:global_attention_prior}
\papertablestyle
\begin{adjustbox}{max width=\linewidth}
\begin{tabular}{@{}l*{10}{c}@{}}
\toprule
\multirow{2}{*}{Method} & \multirow{2}{*}{\shortstack{Retention\\Ratio $\gamma$}} & \multirow{2}{*}{MVBench} & \multirow{2}{*}{LongVB} & \multirow{2}{*}{MLVU} & \multicolumn{4}{c}{VideoMME} & \multicolumn{2}{c}{Mean accuracy} \\
\cmidrule(lr){6-9}\cmidrule(lr){10-11}
 &  &  &  &  & Overall & Short & Medium & Long & Absolute & Relative \\
\midrule
FastVID~\cite{fastvid} & 35\% & \textbf{58.4} & 55.9 & 61.7 & \textbf{58.4} & \textbf{70.8} & \textbf{56.4} & 48.0 & 58.6 & 99.2\% \\
KiToke (Ours) & 35\% & 58.2 & \textbf{57.5} & 62.9 & 58.0 & 70.4 & 54.9 & 48.6 & 59.1 & 100.0\% \\
\textbf{KiToke + global attn.} & 35\% & 58.2 & 57.3 & \textbf{63.7} & 58.3 & 70.7 & 55.0 & \textbf{49.2} & \textbf{59.3} & \textbf{100.3\%} \\
\midrule
FastVID~\cite{fastvid} & 25\% & 58.0 & 56.0 & 61.7 & 58.2 & \textbf{70.3} & \textbf{56.4} & 47.9 & 58.5 & 99.0\% \\
KiToke (Ours) & 25\% & 58.4 & 57.4 & 63.4 & 58.3 & 70.0 & 56.1 & 48.8 & 59.4 & 100.5\% \\
\textbf{KiToke + global attn.} & 25\% & \textbf{58.7} & \textbf{57.6} & \textbf{63.9} & \textbf{58.6} & 70.0 & 56.0 & \textbf{49.7} & \textbf{59.7} & \textbf{101.0\%} \\
\midrule
FastVID~\cite{fastvid} & 15\% & 57.3 & 55.8 & 60.5 & 57.4 & 69.3 & 55.9 & 46.9 & 57.8 & 97.8\% \\
KiToke (Ours) & 15\% & \textbf{57.6} & 56.5 & \textbf{62.5} & 57.7 & 68.8 & 56.0 & \textbf{48.3} & 58.6 & 99.2\% \\
\textbf{KiToke + global attn.} & 15\% & 57.4 & \textbf{56.7} & \textbf{62.5} & \textbf{58.5} & \textbf{69.7} & \textbf{57.5} & 48.2 & \textbf{58.8} & \textbf{99.5\%} \\
\midrule
FastVID~\cite{fastvid} & 10\% & 56.6 & 55.6 & 60.4 & 55.6 & 66.2 & 54.2 & 46.4 & 57.0 & 96.4\% \\
KiToke (Ours) & 10\% & 57.2 & 57.4 & 62.1 & 56.0 & 66.3 & 54.4 & 47.1 & 58.2 & 98.5\% \\
\textbf{KiToke + global attn.} & 10\% & \textbf{57.8} & \textbf{58.2} & \textbf{62.5} & \textbf{56.9} & \textbf{67.0} & \textbf{55.5} & \textbf{48.1} & \textbf{58.9} & \textbf{99.7\%} \\
\midrule
FastVID~\cite{fastvid} & 5\% & 54.8 & 53.7 & 58.3 & 52.6 & 62.6 & 51.0 & 44.3 & 54.8 & 92.7\% \\
KiToke (Ours) & 5\% & 55.2 & 54.9 & 60.8 & 54.1 & 63.0 & 52.2 & \textbf{47.1} & 56.2 & 95.1\% \\
\textbf{KiToke + global attn.} & 5\% & \textbf{55.7} & \textbf{55.1} & \textbf{61.3} & \textbf{54.7} & \textbf{63.8} & \textbf{53.5} & 46.9 & \textbf{56.7} & \textbf{95.9\%} \\
\midrule
FastVID~\cite{fastvid} & 1\% & 45.7 & 46.1 & 51.9 & 46.9 & 52.3 & 46.4 & 41.8 & 47.6 & 80.5\% \\
KiToke (Ours) & 1\% & \textbf{52.5} & \textbf{51.1} & 56.9 & 50.2 & 57.4 & \textbf{49.3} & 43.8 & 52.7 & 89.2\% \\
\textbf{KiToke + global attn.} & 1\% & 52.4 & 50.9 & \textbf{57.1} & \textbf{51.2} & \textbf{59.0} & 49.2 & \textbf{45.4} & \textbf{52.9} & \textbf{89.5\%} \\
\bottomrule
\end{tabular}
\end{adjustbox}
\appendixanalysis{The LLaVA-OneVision results in Table~\ref{tab:global_attention_prior} show gains in four-benchmark mean accuracy at every tested retention ratio. On VideoMME, the prior raises accuracy from 50.2\% to 51.2\% at 1\% retention and from 54.1\% to 54.7\% at 5\%, gains of 1.0 and 0.6 percentage points. By favoring tokens aligned with frame-level semantics, the attention prior adds a saliency signal that complements distribution-aware coverage on these benchmarks.}
\end{table*}

For vision encoders that expose a frame-global pooling token (e.g., a CLS-like token or the SigLIP pooler), we test whether a saliency prior complements distribution-aware selection. For each frame, we obtain the attention score between the frame-global pooling token and each spatial token from the vision encoder, and add this score to the inverse-density score before pivotal sampling. Similar global-attention heuristics are widely used in prior token-compression methods such as VisionZip~\cite{visionzip} and FastVID~\cite{fastvid}. The main comparisons use KiToke without this prior.

\section{Additional Discussion and Theoretical Analysis\label{appendix:limitations}}

\subsection{Query-Agnostic Compression}\label{appendix:query_agnostic_discussion}
KiToke compresses visual tokens before the LLM without using the query, allocating anchors according to full-video pattern prevalence and aggregating features within temporal intervals. We examine these choices in relation to VidCom2~\cite{vidcom2}, FastVID~\cite{fastvid}, FLoC~\cite{floc}, and the query-agnostic pre-LLM component of PruneVID~\cite{prunevid}.

\noindent\textbf{Compression across retention ratios.}
At moderate retention ratios, the token budget can accommodate some redundancy while still retaining useful visual evidence. Even Random preserves 98.3\% of uncompressed accuracy on LLaVA-OneVision at 25\% retention, yet KiToke reaches 100.5\%. KiToke also leads in mean accuracy at 15\% and 25\% retention on LLaVA-Video and Qwen3-VL, supporting the allocation strategy across backbones even when the available budget is relatively generous.

As retention decreases, near-duplicates can occupy a larger share of the budget and crowd out less frequently represented visual cues. Aggregating tokens from semantically different moments can further blur the evidence needed for downstream reasoning. KiToke combines global inverse-density sampling with merging within content-adaptive intervals to address these two sources of information loss.

\subsubsection{Comparison with VidCom2.}

\noindent\textbf{Redundancy signal.}
VidCom2~\cite{vidcom2} scores tokens by comparing them with averaged prototypes, such as global or frame-level means. Although simple and efficient, this approach represents common content through only one or a few reference points. Averaging can therefore obscure local details and motion in highly dynamic videos, limiting the precision of the resulting ``uniqueness'' scores.

Unlike prototype-based scores, KiToke's scores incorporate neighborhood density over the full spatiotemporal token population. Inverse-density weighting down-weights densely represented patterns before sampling, but does not explicitly optimize pairwise diversity of the final anchor set. Tables~\ref{tab:comparison_llava_ov_full}--\ref{tab:comparison_internvl} show that KiToke outperforms VidCom2 across the four evaluated backbones.

\subsubsection{Comparison with PruneVID.}
\noindent\textbf{Factorized redundancy modeling.}
PruneVID's pre-LLM component treats temporal and spatial redundancy separately and operates largely within temporal segments~\cite{prunevid}. This simplifies compression but can miss similar content that recurs at different times and spatial locations, such as recurring backgrounds, repeated shots, and revisited objects. Our density estimates include similar tokens from the full video, allowing recurrence across segment boundaries to affect selection weights. Table~\ref{tab:selection_fixed_merging} isolates this scope: full-video estimation improves mean accuracy over estimation within each interval, with the selection, repair, and merging rules otherwise unchanged.

\noindent\textbf{Temporal segmentation and cost.}
PruneVID clusters pooled frame features with DPC-KNN, using a temporal segment ratio to set the initial cluster count relative to the number of input frames. This establishes the initial granularity, after which feature assignments and continuity refinement determine the final boundaries and segment count.

As described in Section~\ref{sec:method_interval}, KiToke applies fixed thresholds to token-level frame differences and their local increases. Its initial interval count is one plus the number of detected boundaries. When this count exceeds the anchor budget, adjacent intervals are coarsened to meet that budget using the procedure in Appendix~\ref{appendix:empty_intervals}. Figure~\ref{fig:vis_intervals} illustrates the resulting intervals. On LLaVA-OneVision, Table~\ref{tab:efficiency} shows that KiToke achieves higher accuracy and lower compression overhead than baselines that rely on clustering.

\subsubsection{Comparison with FastVID.}
\noindent\textbf{Density and token selection.}
FastVID's density-based selector~\cite{fastvid} combines local density with separation from higher-density tokens to identify representative peaks within selected anchor frames. KiToke uses density in the opposite direction: its selection weights decrease as the number of similar tokens across the full video increases.

Sampling with inverse-density weights keeps tokens in dense regions eligible for selection while giving greater weight to those with fewer substitutes. These scores describe representational redundancy rather than the semantic usefulness of individual tokens. The controlled comparison in Table~\ref{tab:selection_fixed_merging} evaluates the resulting selectors under the same repair and merging pipeline.

\noindent\textbf{Temporal intervals and merging.}
Both FastVID and KiToke merge tokens within temporal intervals, but construct those intervals from different signals. FastVID places boundaries where cosine similarity between adjacent global frame features falls below a threshold. When more boundaries are needed to meet its minimum interval count, it adds the lowest-similarity transitions, so video content determines their locations.

KiToke instead computes change from spatially aligned and nearest-matched token differences, combining the absolute-change criterion on $c_t$ with a threshold on its local increase $\Delta_t$. The additional test can identify a boundary even when $c_t\leq\tau_c$, while the absolute branch still places a boundary at every transition where $c_t>\tau_c$, including a sequence with zero local increases. The component ablation in Table~\ref{tab:token_merge_ablation} shows the benefit of interval-based over global merging.

\subsubsection{Comparison with FLoC.}
FLoC~\cite{floc} uses lazy greedy selection within temporal blocks, with a default block length of 32 frames, to maximize a facility-location objective. This objective sums each original token's cosine similarity to its closest selected representative. Although both FLoC and KiToke seek to represent a video with a limited budget, they account for repetition differently: every occurrence contributes to FLoC's coverage objective, whereas KiToke counterbalances pattern frequency through full-video inverse-density weights. Under the idealized separated-group model, these weights assign equal total sampling mass to groups of different sizes.

The methods also differ in what happens after selection. FLoC retains the selected features directly, while KiToke aggregates unselected features into anchors within temporal intervals using the same inverse-density weights. To isolate the selection rule, Table~\ref{tab:selection_fixed_merging} gives both selectors KiToke's repair and merging pipeline, under which inverse-density selection reaches 52.7\% mean accuracy versus 50.3\% for FLoC.

\subsection{Discussion of Novelty}\label{appendix:novelty}

\subsubsection{Conceptual Novelty} 
\noindent\textbf{\textit{KiToke formulates video token compression as distribution-aware allocation, where a visual pattern's frequency should not automatically determine its share of the token budget.}} Repeated patterns contribute many tokens but often provide overlapping evidence. This motivates a specific design principle: estimate pattern prevalence across the video and use it to counterbalance repetition in both anchor selection and feature aggregation. The formulation determines what redundancy should measure and how that measurement should influence the compressed representation.

\subsubsection{Methodological Novelty}
\noindent\textbf{\textit{Defining redundancy at the appropriate representation level and temporal scope.}} KiToke measures redundancy through the density of the full-video token distribution. Token-level estimation preserves local visual distinctions, while the full-video scope captures repetition across frames and intervals. These choices matter jointly: a pattern can be uncommon within every individual frame yet recur throughout the video. Controlled comparisons with per-frame and per-interval estimation demonstrate that the population used to estimate density materially affects compression quality.

\noindent\textbf{\textit{Connecting reciprocal probabilistic selection to correction for pattern frequency.}} KiToke uses full-video density to allocate sampling mass across visual patterns. Under the analyzed group model, reciprocal weighting cancels pattern multiplicity, whereas uniform weighting favors larger groups and inverse-square weighting favors smaller groups. The robustness result extends this allocation argument to bounded density deviations. Probabilistic selection translates these weights into budget allocation while keeping frequent patterns eligible for selection, with controlled comparisons supporting its advantage over density-proportional and top-$K$ alternatives.

\noindent\textbf{\textit{Maintaining the same distributional principle during feature aggregation.}} Selection determines which tokens become anchors, but aggregation also determines which patterns dominate their features. KiToke therefore uses the same inverse-density scores in both stages, counterbalancing repeated content when allocating anchors and when combining their assigned features. Experiments with identical anchors and assignments isolate the benefit of carrying this principle into aggregation.

\noindent\textbf{\textit{Separating the scope of redundancy estimation from the scope of aggregation.}} KiToke estimates prevalence globally because patterns can recur anywhere in the video, while restricting aggregation to content-adaptive temporal intervals. These intervals are constructed from spatially aligned and nearest-matched token changes, preserving local transition information that global frame representations can obscure. Matched-interval-count comparisons isolate the contribution of this construction beyond the general benefit of temporal locality.

\subsection{Robustness of Group Allocation}\label{appendix:group_allocation}

Proposition~\ref{prop:group_allocation} examines when differences in pattern frequency have little effect on group sampling mass. We consider a fixed set of token embeddings partitioned into $G$ nonempty groups $\mathcal{G}_1,\ldots,\mathcal{G}_G$, with $n_g=|\mathcal{G}_g|$. These groups are used only for analysis, since KiToke computes its scores without clustering. Group membership is disjoint, but similarities between groups may be nonzero.

\noindent\textbf{Kernel conditions.}
Write $K_{ij}=K(\mathbf{x}_i,\mathbf{x}_j)$. Since the Gaussian kernel satisfies $0<K_{ij}\leq1$, sufficient conditions for the proposition are, for every $i\in\mathcal{G}_g$,
\begin{align}
\sum_{j\in\mathcal{G}_g}(1-K_{ij})&\leq\epsilon n_g,\label{eq:within_group_kernel_bound}\\
\sum_{j\notin\mathcal{G}_g}K_{ij}&\leq\eta n_g.\label{eq:cross_group_kernel_bound}
\end{align}
The first condition bounds the loss of within-group similarity, while the second bounds total cross-group similarity relative to group size. Indeed,
\[
D_i=n_g-\sum_{j\in\mathcal{G}_g}(1-K_{ij})+\sum_{j\notin\mathcal{G}_g}K_{ij}
\]
implies $(1-\epsilon)n_g\leq D_i\leq(1+\eta)n_g$. Approximate balance can therefore hold despite imperfect within-group similarity and nonzero cross-group contributions, provided both deviations are small. The cross-group condition bounds the total contribution because even weak individual similarities can accumulate when there are many tokens.

\begin{proof}[Proof of Proposition~\ref{prop:group_allocation}]
Let $W_g=\sum_{i\in\mathcal{G}_g}D_i^{-1}$, $a=(1+\eta)^{-1}$, and $b=(1-\epsilon)^{-1}$. Summing the reciprocal density bounds over the $n_g$ tokens gives
\[
a\leq W_g\leq b.
\]
Because $Q_g=W_g/(W_g+\sum_{h\neq g}W_h)$, it follows that
\[
\frac{a}{a+(G-1)b}\leq Q_g\leq\frac{b}{b+(G-1)a}.
\]
Substituting $\rho=b/a$ gives Equation~\ref{eq:group_allocation_bound}.
\end{proof}

\noindent\textbf{Interpretation and the idealized case.}
The same argument gives $Q_g/Q_h=W_g/W_h\leq\rho$ for any two groups, so their sampling masses remain comparable when $\rho$ is close to one, without explicit dependence on group sizes. In the idealized case of unit similarity within groups and zero similarity between groups, $D_i=n_g$ and $W_g=1$, giving $Q_g=1/G$. By comparison, uniform token weights assign mass $n_g/N$, and density-proportional weights assign $n_g^2/\sum_h n_h^2$. The proposition bounds how far the inverse-density allocation can depart from equal mass for the actual kernel densities at finite $N$.

\noindent\textbf{Expected anchor counts.}
Let $C_g=\sum_{i\in\mathcal{G}_g}I_i$ count sampled anchors before repair, where $I_i$ indicates selection of token $i$. Conditional on the embeddings, pivotal sampling preserves inclusion probabilities, so
\[
\mathbb{E}[C_g]=\sum_{i\in\mathcal{G}_g}\pi_i.
\]
When $\pi_i=Bq_i$, the expectation is $BQ_g$, so multiplying Equation~\ref{eq:group_allocation_bound} by $B$ bounds the expected anchor count. If probabilities saturate, the expectation instead uses the capped and redistributed values in Equation~\ref{eq:inclusion_probabilities}, and interval repair can further change the group counts. The proposition thus describes allocation weights and, under unsaturated sampling, expected counts before repair, without guaranteeing that every group appears in the final anchors.

\subsection{Query-Aware Compression}\label{appendix:query_aware_discussion}
Query-aware methods use the input text to guide visual-token reduction before the LLM, within its layers, or through a hybrid of both stages. KiToke instead allocates a fixed input budget using only video features, which also allows it to precede query-aware pruning in a combined pipeline.

\subsubsection{Instruction-Conditioned Selection}
Query-aware selection can operate before the LLM without extracting its attention maps. CDPruner~\cite{cdpruner} combines pairwise visual-token similarity with token--instruction relevance in a determinantal point process, then greedily selects a subset with high conditional diversity. Both CDPruner and KiToke use relations among visual tokens, but their allocation criteria differ: CDPruner conditions subset diversity on the instruction, while KiToke counterbalances pattern frequency through full-video inverse-density weights.

\subsubsection{Pruning within the LLM}
FastV~\cite{fastv} ranks visual tokens using attention from early LLM layers and removes low-ranked tokens from subsequent layers. SparseVLM~\cite{sparsevlm} selects visually relevant text tokens to score visual tokens, with progressive pruning, adaptive reduction ratios, and token recycling. PyramidDrop~\cite{pyramiddrop} reduces visual tokens at successive stages using attention from the last instruction token. These methods use text--visual interactions formed during LLM computation. Their computational savings depend on both the pruning depth and the retained-token schedule, whereas KiToke reduces the visual input before the first LLM layer.

\subsubsection{Hybrid Video-Compression Pipelines}
Hybrid methods, including HoliTom, FlashVID, and Unified Spatiotemporal Token Compression, combine visual-only compression before the LLM with query-dependent reduction inside it. Their full-pipeline accuracy and cost depend on the token budgets at both stages, including those in earlier layers that are not specified by the final retention ratio.

\noindent\textbf{HoliTom}~\cite{holitom} uses dynamic programming to choose temporal segments that maximize removable redundancy at corresponding spatial positions across consecutive frames. It then applies attention-based spatial selection and density-peak clustering, followed by merging inside the LLM using the last token's attention. Its global optimization determines temporal segmentation, whereas KiToke estimates each token's redundancy against the full video, including recurring content at distant times and spatial locations. These estimates guide global anchor sampling, with feature aggregation confined to temporal intervals.

\noindent\textbf{FlashVID}~\cite{flashvid} combines diversity-based selection calibrated by visual attention and event relevance with tree-based merging across adjacent frames, followed by attention-based pruning inside the LLM. Its selector balances distance from selected tokens with visual importance signals, whereas KiToke derives selection probabilities from each token's prevalence in the full candidate population. Their aggregation rules also differ: FlashVID merges along temporal correspondence trees, while KiToke assigns tokens to anchors within content-adaptive intervals and weights their contributions by inverse density.

\noindent\textbf{Unified Spatiotemporal Compression.}
\citet{unified_stc} maintain a global retention pool, select candidates using visual attention and similarity to already retained tokens, and replenish the pool with clustered features. Their inner-LLM stage then combines text--visual attention and semantic similarity for query-aware merging. Like KiToke, this method allocates a budget across the video. The distinction lies in the allocation rule: their redundancy test compares candidates with the selected pool, whereas KiToke measures pattern prevalence over the full token population before sampling. KiToke uses the resulting inverse-density weights for both anchor selection and feature aggregation.

Our controlled comparison in Table~\ref{tab:allocation_principles} tests pre-LLM allocation with fixed intervals, repair, and merging. Appendix~\ref{appendix:query_composition} evaluates KiToke and FastVID followed by the same FastV stage under a shared 10\%-to-5\% schedule, isolating the choice of pre-LLM compressor within a query-aware pipeline.

\setlength{\dblfloatsep}{8pt plus 2pt minus 2pt}
\setlength{\floatsep}{8pt plus 2pt minus 2pt}
\makeatletter
\setlength{\@dblfpsep}{8pt plus 2pt}
\setlength{\@fpsep}{8pt plus 2pt}
\makeatother
\section{Qualitative Analysis}\label{appendix:qualitative}

\noindent\textbf{Qualitative comparison on temporal interval construction.}
We compare how FastVID~\cite{fastvid} and KiToke divide videos into temporal intervals. FastVID thresholds frame-level differences and enforces a minimum boundary count by adding the highest-difference transitions. KiToke uses token-level change and combines an absolute local-increase test with its absolute-change criterion, as described in Section~\ref{sec:method_interval}.

\begin{figure*}[tp]
  \centering
  \includegraphics[width=\linewidth]{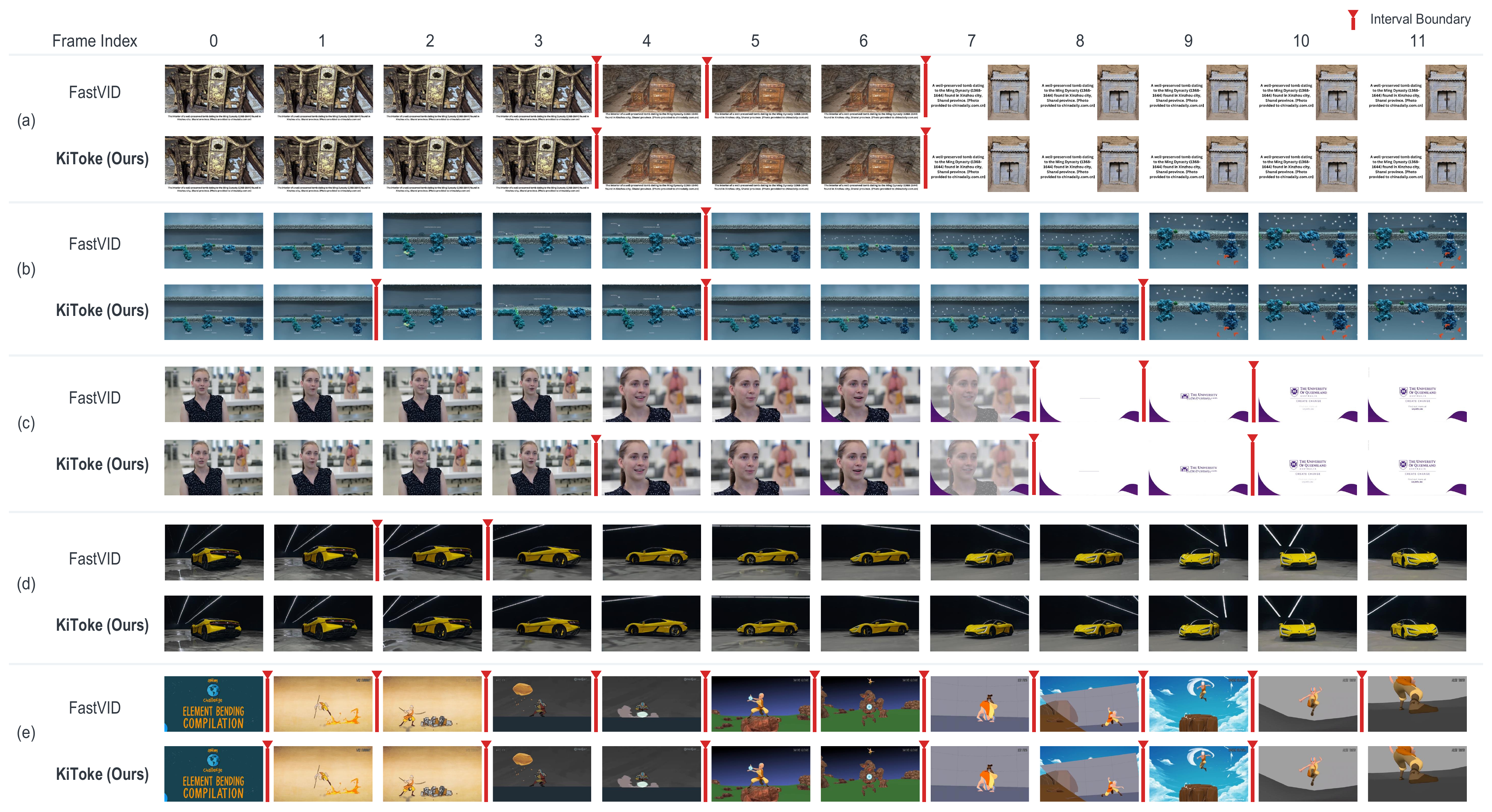}
    \caption{\textbf{Qualitative comparison of temporal interval construction.}
        Each example (a)--(e) shows 12 frames in temporal order, with zero-based display positions 0--11. FastVID and KiToke boundaries are shown in the corresponding rows, marked by red separators and downward triangles. In these examples, KiToke produces boundaries that align more closely with the illustrated content changes than FastVID.}
    \label{fig:vis_intervals}
\end{figure*}

Figure~\ref{fig:vis_intervals} visualizes representative examples. In case (a), FastVID places a boundary between frames 4 and 5 despite minimal visual change because it requires a minimum number of transitions. In cases (b) and (c), KiToke detects subtle changes and places boundaries that more closely follow the visual content. In case (d), KiToke groups the illustrated smooth-motion segment into a single interval. In case (e), FastVID splits nearly every frame, whereas KiToke produces longer intervals over the visually consistent portions shown.

\noindent\textbf{Qualitative comparison on token selection.}
The following five cases compare KiToke's token selection with VidCom2~\cite{vidcom2} and FastVID~\cite{fastvid}, showing how retained visual content relates to the model's answers. The frame panels mark anchor locations rather than reconstructing the merged features, while the accompanying bars show choice-normalized probabilities over A--D.

\noindent\textbf{Choice-normalized probabilities and predictions.}
For LLaVA-OneVision, we take the first assistant-generation step's score vector, \texttt{outputs.scores[0]}, after generation-time logits processing. This position immediately follows the complete multimodal prompt and the \texttt{qwen\_1\_5} conversation template's assistant prefix (\texttt{<|im\_start|>assistant} followed by a newline), before any answer token is generated. The user prompt contains the video, question, and four options, and ends with ``Answer with the option's letter from the given choices directly.''

We score each choice as a single vocabulary entry, obtaining the IDs of the literal uppercase strings \texttt{A}, \texttt{B}, \texttt{C}, and \texttt{D} through the model tokenizer's \texttt{convert\_tokens\_to\_ids} without leading spaces. For their first-step scores $s_c$, we compute
\[
\begin{gathered}
p_c=\frac{\exp(s_c)}{\sum_{d\in\{A,B,C,D\}}\exp(s_d)},\\
c\in\{A,B,C,D\}.
\end{gathered}
\]
These values are next-token probabilities normalized over the four choices, rather than sequence likelihoods of the full option text, and are displayed to three decimal places without confidence calibration. The ``Pred'' label is obtained separately by extracting the first A--D letter from the greedily generated response after removing the evaluator's standard answer prefixes, rather than taking the option with the largest plotted probability.

\begin{figure*}[tp]
  \centering
  \includegraphics[width=\linewidth]{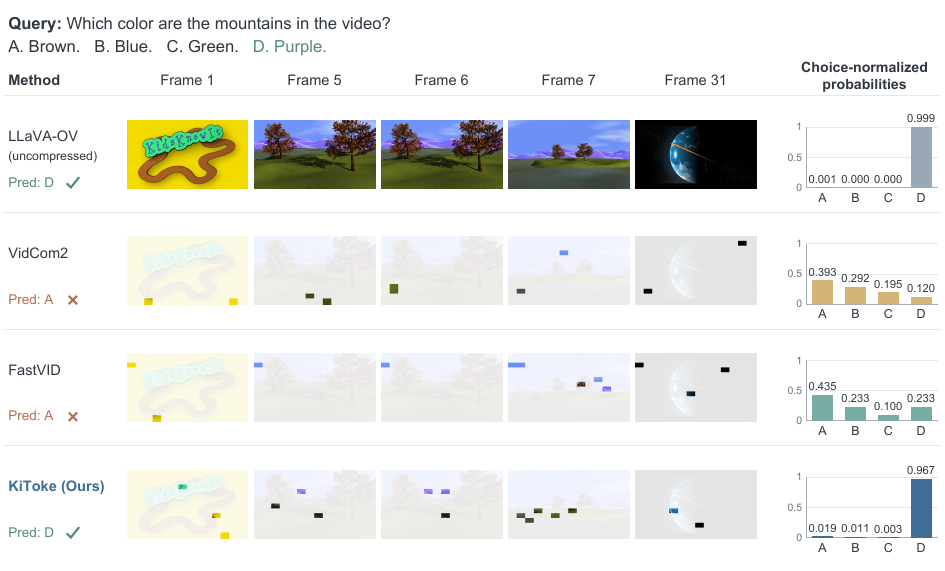}
    \caption{
        \textbf{Qualitative comparison of token selection (case 1).} All methods show the same five frames, labeled with zero-based indices 1, 5, 6, 7, and 31. The bars show choice-normalized probabilities over A--D, and ``Pred'' identifies the option parsed from the generated answer. Compressed methods use $\gamma=1\%$ retention, with LLaVA-OneVision as the uncompressed reference. Anchors are fully opaque, while non-anchor locations are faded and, for KiToke, may contribute through merging. Green marks the ground-truth choice.
    }
    \label{fig:token_selection_case1}
\end{figure*}

\begin{figure*}[tp]
  \centering
  \includegraphics[width=\linewidth]{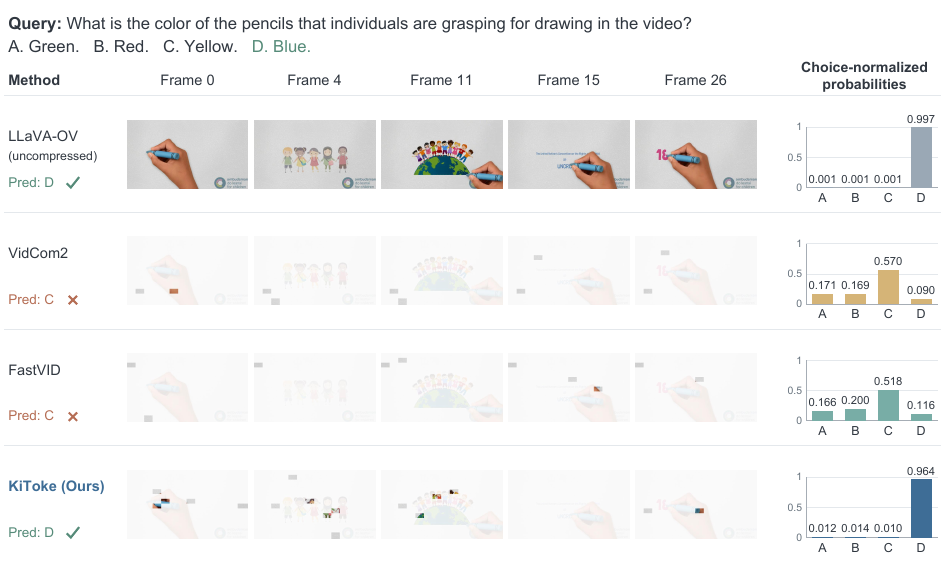}
    \caption{
        \textbf{Qualitative comparison of token selection (case 2).} All methods show the same five frames, labeled with zero-based indices 0, 4, 11, 15, and 26. The bars show choice-normalized probabilities over A--D, and ``Pred'' identifies the option parsed from the generated answer. Compressed methods use $\gamma=1\%$ retention, with LLaVA-OneVision as the uncompressed reference. Anchors are fully opaque, while non-anchor locations are faded and, for KiToke, may contribute through merging. Green marks the ground-truth choice.
    }
    \label{fig:token_selection_case2}
\end{figure*}

Cases~1 and~2 in Figures~\ref{fig:token_selection_case1} and~\ref{fig:token_selection_case2} illustrate successful preservation of query-relevant evidence, including the purple mountains and blue pencils. KiToke retains tokens covering these details, and the model assigns a high probability to the correct choice.

\begin{figure*}[tp]
  \centering
  \includegraphics[width=\linewidth]{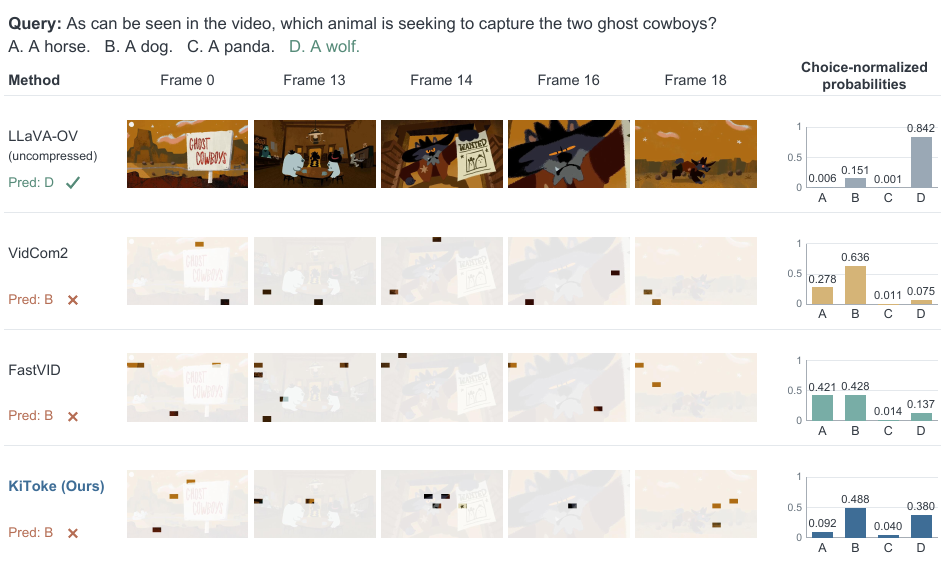}
    \caption{
        \textbf{Qualitative comparison of token selection (case 3).} All methods show the same five frames, labeled with zero-based indices 0, 13, 14, 16, and 18. The bars show choice-normalized probabilities over A--D, and ``Pred'' identifies the option parsed from the generated answer. Compressed methods use $\gamma=1\%$ retention, with LLaVA-OneVision as the uncompressed reference. Anchors are fully opaque, while non-anchor locations are faded and, for KiToke, may contribute through merging. Green marks the ground-truth choice.
    }
    \label{fig:token_selection_case3}
\end{figure*}

Case~3 in Figure~\ref{fig:token_selection_case3} shows that retaining more relevant tokens need not lead to the correct answer. All three compression methods fail: KiToke retains more tokens covering the wolf and assigns a higher probability to the correct option, but still favors dog (B) over wolf (D). The uncompressed LLaVA-OV model assigns probability 0.151 to the incorrect choice B.

\begin{figure*}[tp]
  \centering
  \includegraphics[width=\linewidth]{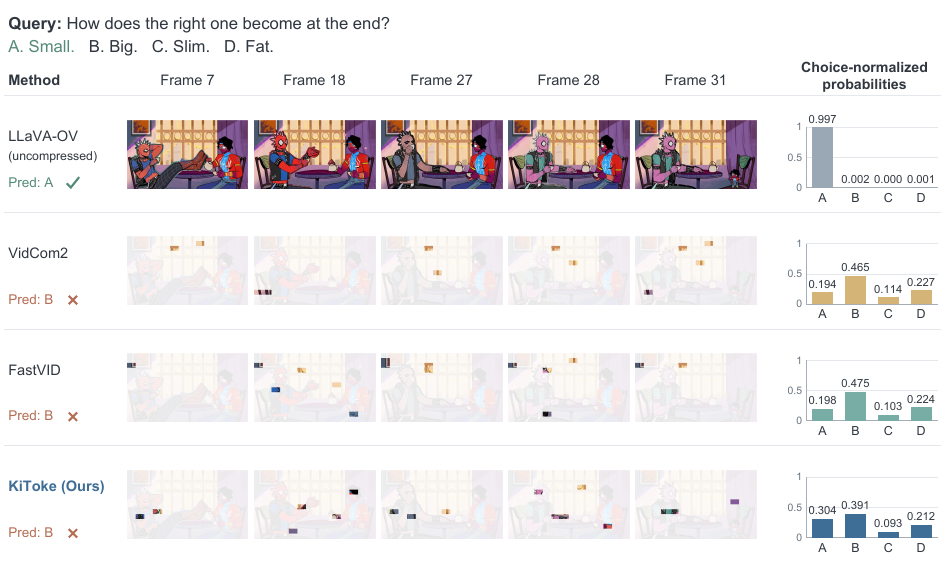}
    \caption{
        \textbf{Qualitative comparison of token selection (case 4).} All methods show the same five frames, labeled with zero-based indices 7, 18, 27, 28, and 31. The bars show choice-normalized probabilities over A--D, and ``Pred'' identifies the option parsed from the generated answer. Compressed methods use $\gamma=1\%$ retention, with LLaVA-OneVision as the uncompressed reference. Anchors are fully opaque, while non-anchor locations are faded and, for KiToke, may contribute through merging. Green marks the ground-truth choice.
    }
    \label{fig:token_selection_case4}
\end{figure*}

Case~4 in Figure~\ref{fig:token_selection_case4} illustrates the difficulty of retaining evidence confined to a small, visually non-salient region under a low token budget. Here the relevant detail appears in a small area of the last frame, which query-agnostic methods can miss unless tokens from that region are retained. KiToke assigns a higher probability to the correct answer than the baselines.

\begin{figure*}[tp]
  \centering
  \includegraphics[width=\linewidth]{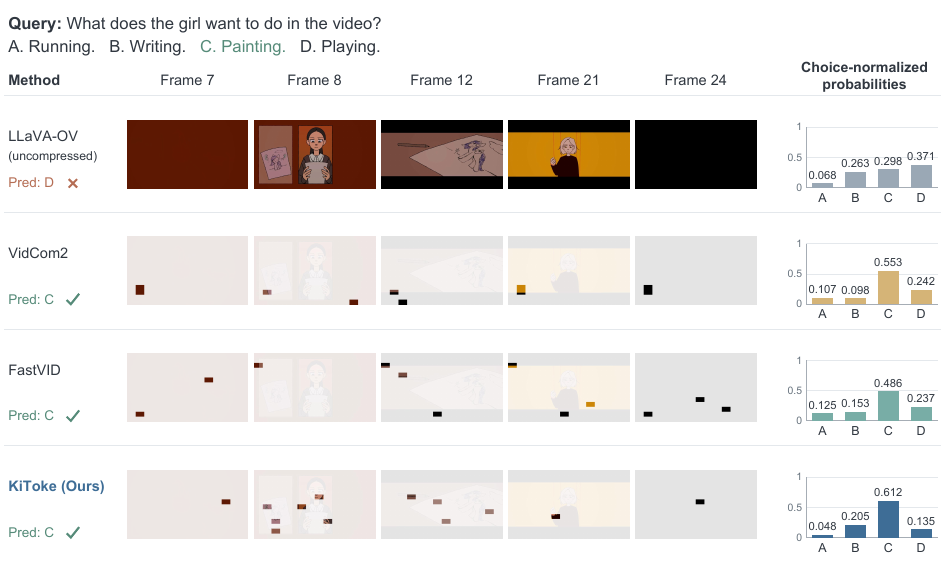}
    \caption{
        \textbf{Qualitative comparison of token selection (case 5).} All methods show the same five frames, labeled with zero-based indices 7, 8, 12, 21, and 24. The bars show choice-normalized probabilities over A--D, and ``Pred'' identifies the option parsed from the generated answer. Compressed methods use $\gamma=1\%$ retention, with LLaVA-OneVision as the uncompressed reference. Anchors are fully opaque, while non-anchor locations are faded and, for KiToke, may contribute through merging. Green marks the ground-truth choice.
    }
    \label{fig:token_selection_case5}
\end{figure*}

Case~5 in Figure~\ref{fig:token_selection_case5} shows the uncompressed model answering incorrectly while the compressed variants answer correctly. One possible explanation is that compression reduces distracting or redundant content, such as background clutter and repeated patterns, clarifying the relevant evidence.

Across these five examples, VidCom2 and FastVID often allocate budget to repeated content or less informative tokens, leaving less room for the evidence needed to answer the questions. The examples illustrate how differences in token allocation can affect which visual details remain available to the model.

KiToke gives more sampling weight to frames containing tokens with fewer similar neighbors across the video. VidCom2 instead sets per-frame budgets using ``token uniqueness'' relative to a global average feature, which can obscure the local details needed to identify decisive evidence. FastVID allocates tokens to each interval in proportion to its frame count. At $1\%$ retention on LLaVA-OV, this allocation can amount to only about two tokens per frame. Although tokens can be redistributed within an interval, the interval's total budget depends on duration rather than full-video pattern prevalence.

\FloatBarrier

\end{document}